\documentclass{article}

\usepackage{amsmath,amsfonts,amsthm,amssymb}
\usepackage{geometry}
\usepackage{graphicx, xcolor} % more modern
\usepackage{subfigure} 

\usepackage{caption}

\usepackage{MnSymbol}
\usepackage{stmaryrd}
\usepackage{wrapfig}
\usepackage{multirow}
\usepackage[mathscr]{euscript}
\usepackage[shortlabels]{enumitem}
\setlist{nolistsep}
\usepackage{array}
\usepackage{diagbox}
\usepackage{float}
\usepackage{framed}

\usepackage[utf8]{inputenc} % allow utf-8 input
\usepackage[T1]{fontenc}    % use 8-bit T1 fonts
\usepackage{hyperref}       % hyperlinks
\usepackage{url}            % simple URL typesetting
\usepackage{booktabs}       % professional-quality tables
\usepackage{amsfonts}       % blackboard math symbols
\usepackage{nicefrac}       % compact symbols for 1/2, etc.
\usepackage{microtype}      % microtypography

\usepackage{verbatim}
\usepackage{algorithm}
\usepackage{algorithmic}
\usepackage{cleveref}

\newtheorem{theorem}{Theorem}%[section]

\DeclareMathOperator{\Prtxt}{Pr}

\newcommand{\prp}[2]{\Prtxt_{#2} \left(#1\right)}

\title{Entropy, concentration, and learning: a statistical mechanics primer}

\author{%
  Akshay Balsubramani \\
  \texttt{akshay@akshay.bio} \\
}

\begin{document}

\maketitle

\begin{abstract}
Artificial intelligence models trained through loss minimization have demonstrated significant success, grounded in principles from fields like information theory and statistical physics. 
This work explores these established connections through the lens of statistical mechanics, starting from first-principles sample concentration behaviors that underpin AI and machine learning. 
Our development of statistical mechanics for modeling highlights the key role of exponential families, and quantities of statistics, physics, and information theory. 
% These foundational ideas are part of a cohesive framework, as we review and refine in a unified manner. 
\end{abstract}

\tableofcontents
\newpage

% \input{SMD_history}
% \chapter{Foundations}
\section{Foundations: entropy, concentration, and learning}

The basic ideas underpinning modern AI and machine learning involve recognition of statistical patterns from sampled training data, which generalize usefully to a test setting. 
The broad probabilistic questions involved here are universal: 

\begin{itemize}
    \item 
    How much can be learned about a probability distribution from a finite sample of it?
    \item 
    What happens when the test-time distribution being evaluated is different from the training distribution?
    \item
    Which patterns and representations usefully generalize to other distributions and datasets? 
\end{itemize}

To begin understanding these basic principles of machine learning, we need to understand concentration of samples from a distribution. 
This understanding has been assembled progressively over time by a succession of quantitative fields encountering these problems, going back to the heart of probability and information theory. 
It was first central to statistical mechanics, a field studying the collective (macroscopic) behavior of large populations of (microscopic) atoms. 

In the late 19th century, as scientists tried to relate properties of bulk matter to its individual atoms, they sought a quantitative theory of this collective behavior. 
Development of the field was stalled at a basic level of understanding for many decades, in which unexplained observations were plentiful and unifying explanations scarce - the problem was daunting because of the huge numbers of microscopic entities involved. 

Progress came from a major insight by Boltzmann. 
When he considered a discretized probability-calculation scenario in trying to develop a molecular theory of gas behavior in 1877 \cite{boltzmann1877beziehung, sharp2015translation}, he launched the field of statistical mechanics \cite{ellis1999theory}. 
These arguments are extremely foundational in an AI context even today, when models learned by loss minimization -- like deep neural networks -- dominate. 

So we first sketch Boltzmann's core statistical mechanics understanding. 
With this entry point, we ultimately describe modern-day (loss-minimization-based) AI modeling in statistical mechanics terms. 
This particular link between modeling and statistical mechanics is foundational, and contains basic insights into learning. 
It allows us to see the essential unity between the basic concepts underlying loss minimization, information theory, and statistical physics.

\subsection{Context among related work}

Starting from Boltzmann, these ideas have been explored in depth and detail, from their discovery in physics in the early 20th century to modern AI/ML. 
Here is a brief discussion to begin. 

These ideas have a long history in the statistical mechanics of Boltzmann and Gibbs. 
Though it was originally developed for physics-centric reasons, early pioneers like Gibbs realized that the math behind it applied to any system of constituent "particles," like a dataset of samples: "The laws of statistical mechanics apply to...systems of any number of degrees of freedom, and are exact" \cite{gibbs1902elementary}. 
In that spirit, very similar ideas would be repeatedly rediscovered and developed in various quantitative fields leading up to modern AI. 

In the postwar years of the 1940s, the basic concepts resurfaced for discrete alphabets with the flowering of probability theory and information theory \cite{shannon1948}. 
The engineering focus of information theory, with real-life consequences built on semiconductor advances in computing, resulted in a different approach and focus to areas such as coding and processing of distributions.

Similar ideas had been developed in statistics as well, through the concept of sufficient statistics \cite{fisher1922mathematical, neyman1936sufficient, koopman1936distributions, Darmois1935, pitman1936sufficient, halmos1949application}. 
The different focus there is reflected in the primary role of exponential families, particularly for easy-to-conceptualize sufficient statistics. 
Interestingly, it took longer to appreciate the connections between this branch of statistics and the ongoing work in information theory. 
Methods of information geometry have more recently developed around some of these ideas. 

The concepts of information theory are applied directly to machine learning in the established subfield of variational inference \cite{wainwright2008graphical, kingma2019introduction, blei2017variational}. 
The focus there has always been on bounding the log-likelihood of the data, accepting maximum-likelihood as a self-evident axiomatic learning principle since early in the field \cite{jordan1999introduction}. 
Information theory tools are used to study, refine, and decompose these bounds. 
We are suggesting a different approach, even though the motivations and some tools are similar. 
By framing the situation differently and looking to explicitly calculate the probabilities, we can develop further principled formulations of learning, and information-theory quantities emerge as a byproduct. 

At present, many AI researchers are aware of these ideas through their appearance in "energy-based models" (EBMs \cite{lecun2006tutorial}), which take the energy function of a Gibbs ensemble as a starting point for modeling, offering interpretable links to such models. 
EBMs have been widely used in modern deep learning, as a class of models for which a wide range of interpretations and tools is available. 

For EBMs, the connections between the learning loss-minimization perspective and the statistical mechanics perspective are applied by formal analogy \cite{lecun2006tutorial, huembeli2022physics}, normally to circumvent a probability-normalization constraint. 
The model class and distribution are typically given a priori, without using the max-entropy perspective or the observation constraints associated with the energy function. 
Related work proceeds in this context \cite{mezard2009information, lamont2019correspondence}, using statistical mechanics notions like energy and temperature as powerful tools to analyze black-box computation \cite{seung1992statistical, saul1996mean}. 
The focus there is substantially different from ours -- the analogy is based on a formal resemblance and not developed in the same way with probability -- and the technical tools used are also different. 

Indeed, the connections between statistical mechanics and statistical inference are not unique. 
There are even more different ways of viewing this, leading to different connections and consequences \cite{watanabe2009algebraic, welling2011bayesian}. 
The quantitative behavior of spin glasses and other less commonly encountered states has turned out to be interesting in this situation \cite{zdeborova2016statistical, bahri2020statistical}.

We will re-develop a statistical mechanics view of the dominant learning principle of loss minimization. 
This parallels the derivation of statistical mechanics, 
showing connections to modeling and prediction which are quite general and powerful.

\subsection{Scope}

From all that we have discussed, the purpose here is to comprehensively lay out the statistical mechanics view of loss minimization from first principles. 
Such scenarios are everywhere in modern AI systems. 
This viewpoint enables us to cross-fertilize ideas between our world of data modeling on one hand, and statistical physics and information theory on the other. 

For the most part, these are well-established concepts and do not require complex technical machinery to handle real-world modeling scenarios. 
So we can prove many general results, in a self-contained way with little needed technically. 
Self-contained derivations of these results are helpful for several reasons:

\begin{itemize}
\item 
The results have well-understood interpretations in the real world, as describing the behavior of bulk observed properties emerging from simply applying some basic underlying principles. 
These interpretations can be translated into data-modeling situations. 

\item 
The methods used for the derivations, from optimization, game theory, and convexity, are familiar tools to AI researchers. 
But in the context of the results in physics and information theory, they lead to sometimes nonstandard proofs and interpretations of known results. 
The proofs' broad scope and applicability, to new problems and loss functions, gives the results significant new power and broadens their applications.

\item
Following developments in AI and modeling, the powerful results of physics and information theory can be extended: to new models, systems with size/regularization which do not occur in observable physics, and more.
\end{itemize}

% \subsection{Information theory preliminaries}

% Information theory describes distributions over some set of possible outcomes $\mathcal{X}$. 
% The basic definition we work from is the cross entropy of a distribution $Q$ with respect to another distribution $P$: 
% $$ \text{H} (P, Q) := \mathbb{E}_{x \sim P} \left[ - \log (Q(x)) \right] $$

% This gives rise to the notions of entropy (the minimum cross entropy) and KL divergence (the regret to the minimum cross entropy):

% \begin{align*}
% \text{H} (P) &:= \text{H} (P, P) = \min_{Q} \text{H} (P, Q) = \mathbb{E}_{x \sim P} \left[ - \log (P(x)) \right]
% \\
% \text{D} (P \mid \mid Q) &:= \text{H} (P, Q) - \text{H} (P) = \mathbb{E}_{x \sim P} \left[ \log \left( \frac{P(x)}{Q(x)} \right) \right]
% \end{align*}

\section{Concentration: Boltzmann's "probability calculation"}
\label{sec:boltzmanncalc}

Start with the first basic question from earlier: 
how much can be learned about a probability distribution from a finite sample of it? 
This question has a precise quantitative answer, which is what Boltzmann found in what he called a "probability calculation" \cite{boltzmann1877beziehung}. 

In probability terms, given a distribution $P$ over a set of outcomes $\mathcal{X}$ (we write $P \in \Delta (\mathcal{X})$), we are looking to understand the consequences of repeatedly sampling from $P$. 
Boltzmann's insight was that if the set of outcomes $\mathcal{X}$ is finite, this question can be answered by explicitly counting the possibilities.  

Suppose $\mathcal{X}$ is finite: $\mathcal{X} = \{x_1, x_2, \dots, x_D \}$. 
We draw $n$ samples from this set independently according to a probability distribution $P = (P_1, \dots, P_D)$, and we observe the frequencies of each outcome. 
Let $n_i$ be the number of times we observe outcome $x_i$, so that $\sum_{i} n_{i} = n$. 
The observed probability of outcome $x_i$ is then $Q_i := n_i / n$, so $Q$ is another probability distribution -- the empirical histogram of the data over $\mathcal{X}$. 

Boltzmann calculated the probability of observing a particular set of frequencies $\{n_1, \dots, n_D\}$ in this situation: 
$$
\begin{aligned}
\frac{1}{n} \log \text{Pr} (x_1 = n_1, \dots, x_D = n_D) 
&= - \text{D} (Q \mid \mid P ) + \frac{1}{2 n} \log \left( \frac{2 \pi n}{ \prod_{i=1}^{D} (2 \pi n P_i) } \right) + \Theta (1 / n^2) \\
\end{aligned}
$$
where the relative entropy (or divergence) of $P$ with respect to $Q$ is $\displaystyle \text{D} (Q \mid \mid P ) := \mathbb{E}_{x \sim Q} \left[ \log \left( \frac{Q(x)}{P(x)} \right) \right]$. 

This is an extremely accurate and powerful approximation for even moderate sample sizes $n$, which tells us the likelihood of observing any specific configuration of outcomes. 

\begin{itemize}
\item
We've calculated the chance of observing a particular set of frequencies $Q$ given $P$. 
If we instead view $Q$ as given, it makes sense to calculate the $P$ which makes the observed $Q$ most likely. 
\item 
The real distribution $P$ only enters the picture through its divergence $- \text{D} (Q \mid \mid P )$ from the observed distribution $Q$. 
\item
This $- \text{D} (Q \mid \mid P )$ is also by far the dominant term, as all the others are $O(1/n)$. 
\end{itemize}

In short, observing the histogram $Q$ alone does not determine $P$, but it does give us enough information to precisely quantify the likelihood of deviations of $Q$ from $P$.

\subsection{Boltzmann's reasoning}

Boltzmann's reasoning is the most direct one even after over a century, and it is worth going over the highlights. 

Boltzmann reduced the problem to essentially a generalized balls-in-bins problem by discretizing the probability space, and discretizing quanta of probability (at resolution $\frac{1}{n}$). 
The calculation is simply a matter of accounting for the differently weighted "bins" (outcomes), and the combinatorially many ways of throwing "balls" (quanta of probability) into them. 
\begin{align*}
\text{Pr} (x_1 = n_1, \dots, x_D = n_D) 
&= \frac{n!}{n_1! \cdots n_D!} P_1^{n_1} \cdots P_D^{n_D} \\
&= \frac{n!}{n_1! \cdots n_D!} \exp \left( n_1 \log P_1 + \cdots + n_D \log P_D \right) \\
&= \frac{n!}{(n Q_1)! \cdots (n Q_D)!} \exp \left( n \sum_{i=1}^{D} Q_i \log P_i \right) \\
&= \frac{n!}{(n Q_1)! \cdots (n Q_D)!} \exp \left( - n \text{H} (Q, P) \right)
\end{align*}

where the cross entropy of $P$ to $Q$ is $\displaystyle \text{H} (Q, P) := \mathbb{E}_{x \sim Q} \left[ - \log \left( P(x) \right) \right]$. 

We can rewrite the multinomial coefficient using Stirling's approximation ($ \log n! \approx n \log n - n + \frac{1}{2} \log (2 \pi n) + \Theta (1/n) $).

\begin{align*}
\log &\left( \frac{n!}{(n Q_1)! \cdots (n Q_D)!} \right) 
= \log (n!) - \sum_{i=1}^{D} \log ((n Q_i)!) \\
&= n \log n - n + \frac{1}{2} \log (2 \pi n) - \sum_{i=1}^{D} \left[ n Q_i \log n Q_i - n Q_i + \frac{1}{2} \log (2 \pi n Q_i) \right] + \Theta (1/n) \\
&= n \log n + \frac{1}{2} \log (2 \pi n) - n \sum_{i=1}^{D} Q_i \log n Q_i - \frac{1}{2} \sum_{i=1}^{D} \log (2 \pi n Q_i) + \Theta (1/n) \\
&= n \text{H} (Q) + \frac{1}{2} \log (2 \pi n) - \frac{1}{2} \sum_{i=1}^{D} \log (2 \pi n Q_i) + \Theta (1/n)
\end{align*}
where the entropy of $P$ is $\displaystyle \text{H} (P) := \text{H} (P, P) = \min_{Q} \text{H} (P, Q) = \mathbb{E}_{x \sim P} \left[ - \log (P(x)) \right]$. 

Therefore, the probability of observing the frequencies $n_1, \dots, n_D$ is 
$$
\begin{aligned}
\text{Pr} (x_1 = n_1, \dots, x_D = n_D) 
&= \exp \left( - n \sum_{i=1}^{D} Q_i \log \frac{Q_i}{P_i} + \frac{1}{2} \log \left( \frac{2 \pi n}{ \prod_{i=1}^{D} (2 \pi n Q_i) } \right) + \Theta (1/n) \right) \\
\end{aligned}
$$

Restating this gives the result.

\subsection{Consequences}

What Boltzmann called his ``probability calculations" \cite{boltzmann1877beziehung} launched the field of statistical mechanics, 
and inspired the field of information theory decades later. 
This is because discretizing the space is a fully general technique, 
with all the essential elements used to study concentration and collective behavior in statistical mechanics. 
The major quantities of information theory -- entropy $\text{H} (P)$, cross entropy $\text{H} (P, Q)$, and relative entropy (divergence) $\text{D} (P \mid \mid Q)$ -- all emerge directly from the calculation, as the evident quantities of interest. 

The calculation shows the degeneracy in observing the histogram $Q$ -- the "macrostate" of the $n$-sample dataset -- from a particular "microstate," i.e. the individual outcomes of each of the $n$ samples. 
This was the idea that allowed physicists to quantify observable bulk properties of matter (macrostates) from unobservable configurations of each of its atoms (microstates). 

In AI and data science, the system being studied (the "matter") is a dataset comprising $n$ examples, whose state is the microstate. 
And the macrostate consists of our coarse-grained observations about the dataset, as we develop more in the following sections.

\section{Enter entropy}

In this calculation, the log-multinomial coefficient $\log \left( \frac{n!}{(n P_1)! \cdots (n P_D)!} \right)$ is $\approx n \text{H} (P)$, with the approximation being very accurate for even moderate $n$. 
In fact, this is the only approximation that we have made in the calculation. 
How accurate is it?

\begin{figure}
\begin{center}
    \scalebox{1}{
        \includegraphics[width=145mm, height=80mm]{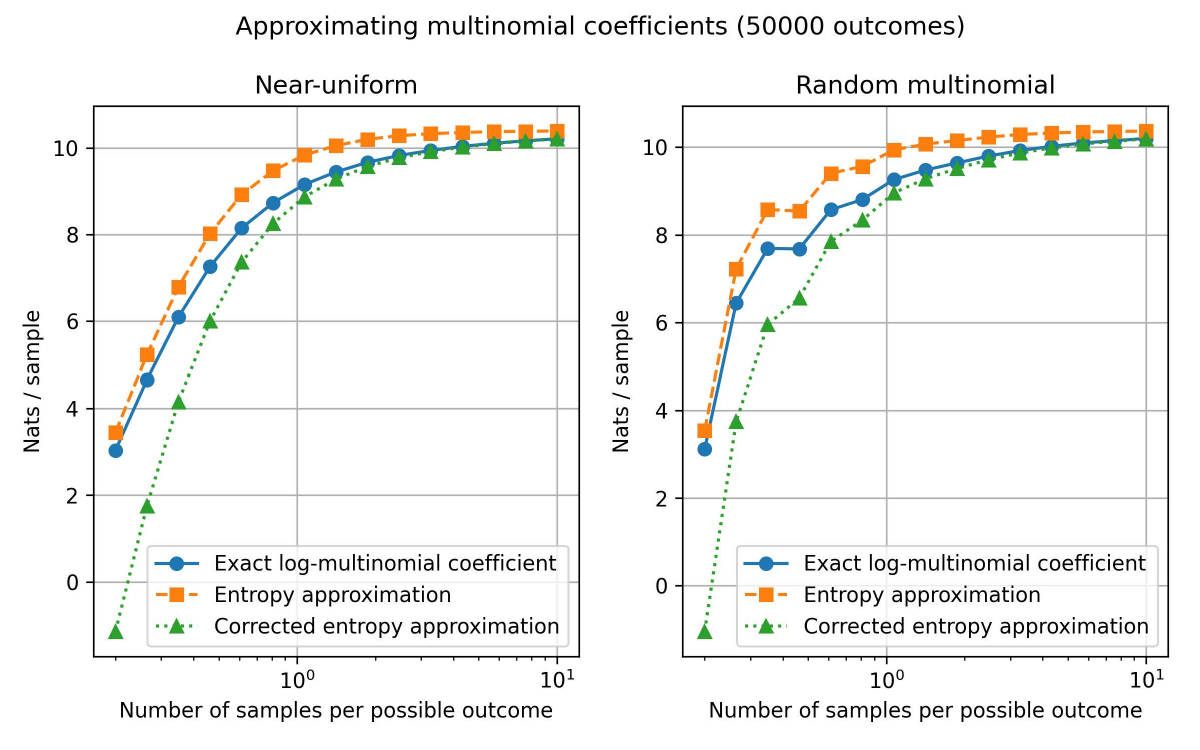}}
\end{center}
\caption{Approximating entropy for different distributions, varying $n$, for $D = 50000$. At left is for a near-uniform distribution $P$ (parameters sampled according to a Dirichlet(1)). At right is for $P$ sampled uniformly over the positive orthant, for which much more information is contained per sample. Note that even for moderate $n$ ($n / D \geq ~5$) -- the relative error is just a small constant. Meanwhile, the first-order correction $\frac{1}{2} \log \left( \frac{2 \pi n}{ \prod_{i=1}^{D} (2 \pi n P_i) } \right)$ makes the estimate significantly more accurate. }
\label{fig:logmultapprox}
\end{figure}

We can quantify the relative probability, i.e. the exp-difference between the log-multinomial coefficient and $n \text{H} (P)$ (Fig. \ref{fig:logmultapprox}). 
This shows that the approximation is very accurate (and that extreme accuracy is achieved when the first-order correction is applied), even for distributions over $50,000$ outcomes, comparable to modern LLM token vocabularies. 

This multinomial coefficient is the number of ways to get the same observed macrostate (the histogram $Q$) from particular microstates (the individual values of all the examples in the dataset). 
In other words, microstates can be counted in terms of the entropy of the observed macrostate. 

Putting all this together, we arrive at some powerful insights.

\begin{itemize}
\item
Entropy is a measure of the microscopic multiplicity, or degeneracy, associated with a set of limited macroscopic/average observations into the underlying microstate. 
High-entropy configurations are exponentially more likely than other configurations - they dominate observed configurations for large $n$. 
\item 
Therefore, the macrostate maximizing entropy is the "most likely" state. Entropy maximization accounts for the many microstates that are consistent with a set of macroscopic observations. 
\end{itemize}

In the macroscopically sized samples of atoms we observe in everyday life, our observations are almost deterministic, even of highly disordered systems like gaseous and liquid matter. 
This is because they are made of moles of individual constituents, and our handful of bulk properties observed about them corresponds to only a handful of constraints. 

In the statistical mechanical view of AI, when trying to minimize loss over datasets, we can similarly say that the entropy of the learned distribution tends to be nearly maximal. 
(We later prove this, in a very general sense.)

What we've described, with a discrete "alphabet" of possible states, is the concept called the \emph{asymptotic equipartition property} in information theory -- high-entropy sequences occur with much greater multiplicity than low-entropy ones \cite{CT06}. 
Boltzmann's "probability calculation" shows exactly why, with entropy emerging as a measure of (log) multiplicity.

\section{Learning: generalizing Boltzmann's scenario}
\label{sec:generalizingboltzmann}

Until this point, we have followed the clarifying insight of Boltzmann in looking only at discrete outcome spaces $\mathcal{X}$, which has allowed us to do explicit probability calculations. 
Only a finite set of outcomes can happen there -- 
in the AI context, it is like requiring each training sample to be a member of some discrete space. 
This is itself very useful in practice for models over text and similar discrete spaces. 

However, practical modern scenarios are also full of continuous spaces like $\mathcal{X} = \mathbb{R}^{d}$. 
This makes it impossible to assign probabilities pointwise and count their combinatorics discretely. 
How, then, can we quantify observations about the distribution $P$? 

First, we no longer observe a histogram $Q$ over the discrete $\mathcal{X}$, as Boltzmann did. 
Instead, the natural extension of the observed $Q$ is the "empirical measure" $\hat{P}_n$, which puts weight on any event according to the event's frequency over the $n$ samples.\footnote{More precisely, $\hat{P}_{n} (E) := \frac{1}{n} \sum_{i=1}^{n} \textbf{1} ( x_i \in E )$ is the empirical measure associated with the $n$-sample $(x_1, \dots, x_n)$.}

% There is a coarse-graining in what we observe, as we do not observe the full distribution $P$. 

% Then the empirical measure $\hat{P}_n$ is a sufficient statistic for the distribution $P$ if the feature functions are rich enough. 

% This is where the machine learning perspective shows its power. 

Also, each observation corresponds to a function of the outcome $f_i (x)$, which associates a real number to any outcome $x \in \mathcal{X}$. 
In AI / machine learning, this is a \textit{feature function} -- we observe $d$ of them.
For any feature function $f_i$, our observation over the data is the empirical average over the sampling distribution $\hat{P}_n$: $\mathbb{E}_{x \sim \hat{P}_n} [f_i (x)]$. 
We observe $\forall i$ that $\mathbb{E}_{x \sim \hat{P}_n} [ f_i(x) ] = \alpha_i $ for some $\alpha_i \in \mathbb{R}$, i.e. that $\hat{P}_n \in \mathcal{A}$, where 
\begin{align*}
\mathcal{A} := \left\{ P \in \Delta (\mathcal{X}): \mathbb{E}_{x \sim P} [ f_i(x) ] = \alpha_i \;\forall i = 1, \dots, d \right\}
\end{align*}

With these concepts in mind, the situation is a natural extension of Boltzmann's calculations arising above for discrete $\mathcal{X}$. 
The discrete Boltzmann scenario can be fully generalized, in a beautiful way that retains all the insights observed before.

\section{Concentration: general "probability calculations"}
\label{sec:genprobcalc}

The quantity we're interested in is still the probability of seeing the observations, just like in Boltzmann's case. 
In our new scenario with the generalized notation, this is 
$$ \text{Pr} \left( \hat{P}_n \in \mathcal{A} \right) $$

The behavior of this probability -- or rather the normalized log probability of the observation $ \frac{1}{n} \log \text{Pr} \left( \hat{P}_n \in \mathcal{A} \right) $ -- has been extensively bounded  by a series of results (the theory of \emph{large deviations}).

\subsection{The general calculation}
\label{sec:subgenprobcalc}

Now we are in a new setting of general $\mathcal{X}$, with the concepts of $\hat{P}_{n}, \{ f_i (x) \}_{i=1}^{d} , \mathcal{A}$. 
For the first time we encounter an extremely important definition.

\subsubsection{Information projection}

The information projection of $P$ on $\mathcal{A}$ is the distribution $P_{\mathcal{A}}^{*} \in \mathcal{A}$ that is closest to $P$, according to divergence with $P$ used as a prior. 
\begin{align*}
P_{\mathcal{A}}^{*} := \arg\min_{Q \in \mathcal{A}} \text{D} (Q \mid\mid P )
\end{align*}

This is almost always the target distribution that we try to learn, changing $\mathcal{X}, \mathcal{A}$ to suit the situation. 
$P_{\mathcal{A}}^{*}$ is desirable for modeling data under observations $\mathcal{A}$ -- in fact, it is essentially universal and unique because of its many favorable and fully general properties:

\begin{itemize}
\item
\textbf{Admissibility}: $P_{\mathcal{A}}^{*}$ meets the constraints $\mathcal{A}$. 
\item 
\textbf{Highest probability / "likelihood"}: $P_{\mathcal{A}}^{*}$ is the distribution that is most likely to have generated the observed data. 
\item 
\textbf{Axiomatic justification}: $P_{\mathcal{A}}^{*}$ is essentially the only distribution that could be generating the data, in which measuring the data with the given features does not lose relevant information. 
\item 
\textbf{Robustness}: $P_{\mathcal{A}}^{*}$ is the most robust distribution to predict with, given the expected feature values across the dataset. 
\item
\textbf{Convenience}: It has many favorable properties for approximation of data, and for being conveniently learnable. 
Learning problems are typically convex, and the error decomposes readily in convenient ways. 
\end{itemize}

We discuss all these properties in \Cref{sec:seclearning}.

\subsubsection{The result}

For any set $\mathcal{A}$ that is an intersection of expected-value constraints as defined in \Cref{sec:generalizingboltzmann}, 
we can rewrite the probability of observing $\hat{P}_n$ in $\mathcal{A}$ as 
\begin{align*}
\frac{1}{n} \log \text{Pr} \left( \hat{P}_n \in \mathcal{A} \right) 
= - \text{D} (P_{\mathcal{A}}^{*} \mid\mid P ) - \frac{1}{n} \text{D} ( \mu_{\mathcal{A}} \mid\mid P_{\mathcal{A}}^{*n} )
\end{align*}
where the conditional distribution $\mu_{\mathcal{A}}$ is the data $P^{n}$ conditioned on the empirical measure falling in $\mathcal{A}$. \footnote{More completely, $\displaystyle \mu_{\mathcal{A}} (y) := \text{Pr}_{Z \sim P^{n}} \left( Z = y \mid \hat{P}_{n} \in \mathcal{A} \right)$, for $y \in \mathcal{X}^n$. }
We emphasize that this is an identity, not a bound -- so all concentration bounds in this setting are essentially approximating this one. 
It can be proved in great generality using only a short argument with basic techniques \cite{balsubramani2020sharp}. 

The identity, only involving the fundamental quantities related to $\mathcal{A}$, is bounded and approximated by a line of "Sanov-type" results, following the main result of Sanov \cite{Sanov57}: 

\begin{theorem}[Sanov's Theorem]
\label{thm:asymp-sanov}
\begin{align*}
\lim_{n \to \infty} \; \frac{1}{n} \log \prp{ \hat{P}_n \in \mathcal{A}}{} 
= - \min_{Q \in \mathcal{A}} \text{D} (Q \mid\mid P)
= - \text{D} (P_{\mathcal{A}}^{*} \mid\mid P)
\end{align*}
\end{theorem}

This is the appropriate way of generalizing the probability calculation for finite $\mathcal{X}$. 
Again, the message is that the dominant term in the log-probability is $- \text{D} (P_{\mathcal{A}}^{*} \mid\mid P)$; note the similarities with Boltzmann's result for discrete $\mathcal{X}$.

\subsection{Discussion: the nature of concentration in $\mathcal{A}$}

Sanov's theorem is a very general result that is well known to describe many concentration phenomena and subsume many concentration inequalities. 
The relative entropy term ($- \text{D} (P_{\mathcal{A}}^{*} \mid\mid P )$) on the right-hand side typically dominates, and is the pivotal quantity studied by the theory of large deviations \cite{ellis1999theory}. 

As such a general statement, Sanov's theorem too has a long history out of the immediate scope here, dating back to the origins of information theory on bit-strings \cite{CT06}. 
The modern literature \cite{touchette2009large, dembozeitouni2009large} develops many other consequences and ideas of this kind.

\subsubsection{Gibbs conditioning principle}

In conjunction with the finite-$n$ identity $ \frac{1}{n} \log \text{Pr} \left( \hat{P}_n \in \mathcal{A} \right) = - \text{D} (P_{\mathcal{A}}^{*} \mid\mid P ) - \frac{1}{n} \text{D} ( \mu_{\mathcal{A}} \mid\mid P_{\mathcal{A}}^{*n} )$, 
we see that 
\begin{align*}
\lim_{n \to \infty} \text{D} ( \mu_{\mathcal{A}} \mid\mid P_{\mathcal{A}}^{*n} ) = 0
\end{align*}

(This fact can be shown independently \cite{van1981maximum, tjur1974conditional, csiszar1984sanov}. 
If it is taken alternatively as a starting point, it can be combined with the finite-$n$ identity to prove Sanov's theorem!)

So the conditional distribution  $\mu_{\mathcal{A}}$ behaves like $P_{\mathcal{A}}^{*n}$, as if it were $n$ i.i.d. samples from $P_{\mathcal{A}}^{*}$. 
This originates from the roots of statistical mechanics over a century ago \cite{gibbs1902elementary}, often known as the Gibbs conditioning principle \cite{leonard2002extension}. 
This is quite a profound idea -- the approximating distribution $P_{\mathcal{A}}^{*n}$ completely removes the interdependences between the $n$ samples of conditioning in $\mu_{\mathcal{A}}$. 

In effect, we can "pretend" the data are i.i.d. generated from $P_{\mathcal{A}}^{*}$, a topic we return to later in describing the properties of exponential families.

\subsubsection{The impact of further information}

We can actually go much further in outlining the role of $P_{\mathcal{A}}^{*}$ in the tail probability of additional information, $\mathcal{B} \subseteq \mathcal{A}$. 

It is possible to prove a fully general formula for the relative probability of a subset $\mathcal{B} \subseteq \mathcal{A}$ \cite{balsubramani2020sharp}: 
\begin{align*}
\log \text{Pr} \left( \hat{P}_n \in \mathcal{B} \mid \hat{P}_n \in \mathcal{A} \right) 
= - \left( \text{D} ( \mu_{\mathcal{B}} \mid\mid P_{\mathcal{A}}^{*n} ) - \text{D} ( \mu_{\mathcal{A}} \mid\mid P_{\mathcal{A}}^{*n} ) \right)
\end{align*}

This shows the role of the $n$-sample exponential family $P_{\mathcal{A}}^{*n}$ in measuring the impact of further information $\mathcal{B} \setminus \mathcal{A}$ on the log-probability of an $n$-sample from $P$. 
The added information affects the probability in a way that depends on how much it affects the regret of predicting with $P_{\mathcal{A}}^{*n}$. 
Among product distributions when $\mathcal{A}$ is known, this regret is minimized by predicting with $P_{\mathcal{A}}^{*n}$; augmenting the knowledge to $\mathcal{B} \subseteq \mathcal{A}$ only increases the regret.

\section{Entropy: priors and perturbations}

We have seen that maximum-entropy microstates have the highest multiplicity under the macroscopic observations, when the outcome space $\mathcal{X}$ is discrete. 
What is the probability of observing some other microstate of the system (dataset), even if it's a perturbation away from having maximal entropy?

The modern treatment of such "fluctuations" constitutes some of the foundations of statistical mechanics (see \cite{schrodinger1948statistical, landaulifshitz}), typically attributed to Einstein \cite{einstein1910theorie} and by him to Boltzmann. 
The same ideas are extremely useful in motivating the meaning of entropy.

\subsection{The prior as a carrier measure}

First, we need to discuss the role of the base "carrier" measure over the microstate space $\mathcal{X}$. 
This is something we specify here, equivalent to the prior we put over our inferences. 
As \cite{grunwald2007minimum} says, the carrier measure "...represents the symmetries of the problem, which amounts to determining how outcomes should be counted." 
Reflecting this, there are more sophisticated and general group-theoretic arguments for how to encode ignorance in different spaces \cite{jaynes1968prior}. 

Ultimately, this is a free and arbitrary choice, dictated by philosophy \cite{jaynes1986monkeys} and the practical situation at hand. 
The choice matters to the performance of any downstream inference, in a way that's well-studied by information theory -- the more the choice reflects the test-time reality, the better it performs at test-time inference. 

In learning, this corresponds to choosing a prior, which evidently affects the loss. 
The learning literature is full of development of the relationship between the loss (often some version of log loss), the regularization, and the prior \cite{murphy2012machine}. 
Learning scenarios therefore use the prior flexibly, defining it differently for each situation. 
But it's useful to discuss a basic default option first, which shows how the prior determines "how outcomes should be counted."

\subsection{A uniform prior: entropy counts probabilities}
\label{sec:unifprior}

Let's look at the consequences of making this measure uniform over known outcomes, as an expression of our prior indifference between them. 

Writing this prior as $P_{0}$, we can denote the resulting empirical measure and information projection as $\hat{U}_{n}$ and $U_{\mathcal{A}}^{*}$ respectively, and write using Sanov's theorem: 
\begin{align}
\frac{1}{n} \log \text{Pr} \left( \hat{U}_n \in \mathcal{A} \right) 
&= - \text{D} (U_{\mathcal{A}}^{*} \mid \mid P_{0}) - \text{D} (\mu_{\mathcal{A}} \mid \mid U_{\mathcal{A}}^{* n}) \\
&\leq - \text{D} (U_{\mathcal{A}}^{*} \mid \mid P_{0})
= \text{H} (U_{\mathcal{A}}^{*}) - \text{H} (U_{\mathcal{A}}^{*}, P_{0})
\end{align}

In this case, $\text{H} (x, P_{0})$ is the same for all $x$; call this value $\ell (P_{0})$. 
Then we have concluded that: 
$$
\text{Pr} \left( \hat{U}_n \in \mathcal{A} \right) 
\leq \frac{ e^{n \text{H} (U_{\mathcal{A}}^{*})}}{ e^{n \ell (P_{0})}} 
\propto \exp \left( n \text{H} (U_{\mathcal{A}}^{*}) \right)
$$

So if we implicitly assume that the carrier measure over the data is uniform over the outcome space $\mathcal{X}$, \textbf{the log-probability of any macrostate $\mathcal{A}$ is determined by the entropy $H (U_{\mathcal{A}}^{*})$.} 

This was a very early discovery of Boltzmann in the context of physics, where the uniform carrier measure is typically justified by Liouville's theorem characterizing how physical systems evolve in "phase space." 
Boltzmann was so pleased with it that he had it inscribed on his tombstone. 
Via Einstein \cite{einstein1910theorie}, it made its way into standard treatments of statistical mechanics \cite{schrodinger1948statistical, landaulifshitz}.

In learning scenarios, uniformity over the outcome space makes sense as well. 
It is often a default choice because of computational convenience, and a desire to avoid ruling out any regions of the outcome space -- with high enough $n$, commonly used learning procedures converge to the correct answer regardless of prior. 
And when the $n$ constituents are i.i.d. sampled data, uniform weights are extremely natural. 
(However, learning scenarios also suggest non-uniform prior distributions, which generalize the applications of statistical mechanics. )

In short, \textbf{both statistical mechanics and learning scenarios use a uniform prior in certain situations, for different reasons}.

\subsection{Fluctuations}
\label{sec:fluctuationsgaussian}

Since $\mathcal{A}$ does not constrain any feature directly but only its average, we expect fluctuations in the observed features, and can precisely quantify them. 
In statistical physics, this has been studied for a long time on a basic level \cite{landaulifshitz, kardar2007statistical}, in the context of fluctuations in observed energy and other quantities.

The idea is that any system comprised of many separately observed units shows it upon even a bulk observation. 
The many teeming units lead to predictable probabilistic behavior, as predicted by statistical mechanics. 
When $n$ is large, the fluctuations are negligible, and for gigantic $n \sim 10^{23}$ as in physically observable systems, the fluctuations can be unobservably small. 
Physics handles this situation through many approximations, which only hold in the large-$n$ limit. 
Statistical physics for small-$n$ is typically confined to relatively exotic systems in the observable world. 
But learning scenarios are very different -- $n$ could be any size.

There is no exact parallel between the loss-minimization / probability-theory scenario and the energy-based one of statistical physics. 
Energy is an observation which happens to have a privileged status in physics, compared to other observations like volume and particle number. 
In our view of statistical mechanics, the observations are the features $f_{i}$, which are all kept on the same footing. 
Each feature function corresponds to just one constraint, just like energy does -- so any feature could be considered to play the role of energy. 

In general, feature fluctuations happen with frequencies governed approximately by $P_{\mathcal{A}}^{*}$ -- an extreme case of this is the uniform prior, as we have shown in \Cref{sec:unifprior}. 
The exact picture is given by $P_{\mathcal{A}}^{*}$, and makes exponential families important in general. 
`\section{Learning: a prescription}
\label{sec:seclearning}

In all this -- the core problem of concentration, and the natural role of entropy in counting combinations of microstates -- we're motivated by learning the data distribution $P$ from observations $Q$. 

It's interesting to highlight some very differently motivated ways to proceed with this learning problem. 
It turns out that they are all equivalent, so they provide complementary perspectives that we'll describe, all of which amount to maximizing entropy in the correct context. 

The maximum-entropy method can be viewed directly as a prescription for learning from data, in a straightforward scenario we've outlined previously in \Cref{sec:generalizingboltzmann}. 
To summarize the scenario: 

\begin{itemize}
\item 
We know the data space $\mathcal{X}$, and we have a set of feature functions $f_i : \mathcal{X} \to \mathbb{R}$, $i = 1, \ldots, d$ that we can observe over $\mathcal{X}$. 

\item 
We sample $n$ elements from $\mathcal{X}$ using an unknown distribution $P$, giving an empirical measure $\hat{P}_{n}$. 

\item
We observe the expected values of these features $f_i$ over some data distribution $\hat{P}_{n}$, $\mathbb{E}_{x \sim \hat{P}_{n}} [f_i (x)] = \alpha_i$, $i = 1, \ldots, d$. 
So $\hat{P}_{n} \in \mathcal{A}$, where again remember that 
\begin{align*}
\mathcal{A} := \left\{ P \in \Delta (\mathcal{X}): \mathbb{E}_{x \sim P} [ f_i(x) ] = \alpha_i \;\forall i = 1, \dots, d \right\}
\end{align*}
\end{itemize}

To model the data distribution $P$ in this scenario given the observations $\hat{P}_{n} \in \mathcal{A}$, we can view the extended Boltzmann calculation above in some more general ways for learning.

\subsection{Some equivalent perspectives}
\label{sec:subequivlearning}

It's useful to show some completely complementary perspectives on learning. 
They inevitably all suggest the same common inference method, but they seem on the surface to be very different from each other. 
Each illuminates a different aspect of learning, as we discuss in \Cref{sec:subequivalencelearning}.

\subsubsection{Prescription I: minimizing log loss in a model class}

Start with a familiar justification for learning: 

\textit{Predict with a distribution $Q$ that minimizes the log loss to the data $\hat{P}_{n}$ (cross-entropy $\text{H} (\hat{P}_{n}, Q)$). }

Since $\hat{P}_{n}$ is the empirical distribution of the data we have, we are trying to learn the distribution $Q$ that best fits the data. 
If the data distribution $\hat{P}_{n}$ is known to match $P$ exactly, then the learning problem is: 
\begin{align*}
\min_{Q \in \Delta(\mathcal{X})} \text{H} (\hat{P}_{n}, Q)
\end{align*}

The naive solution here is a trivial one: $Q = \hat{P}_{n}$, recapitulating the empirical observations perfectly. 
Why is this trivial?

\begin{itemize}
\item 
\textbf{Generalization}: In real situations, where $\mathcal{X}$ is continuous or high-dimensional, we cannot expect $\hat{P}_n$ to totally generalize to $P$ -- no two samples $\hat{P}_{n}$ are exactly alike.
This is a common and general situation we are faced with in machine learning. 
\item 
\textbf{Regularization}: The typical solution is to guide the modeling by restricting $Q$ within $\Delta (\mathcal{X})$, with various powerful model classes available for $Q$ like deep architectures. 
These build inductive bias into the modeling, in a way which we wish will generalize to test sets. 
\end{itemize}

So in this formulation, something has to change; typically, we introduce some model assumptions on $Q$. 
By restricting $Q$ to a still-expressive family of modeling distributions, we can hope that the eventually learned $Q$ will generalize past the sample $\hat{P}_{n}$ to $P$. 

We will consider a common and universally used type of modeling distribution, where $Q$ is a member of an exponential family, an easy-to-work-with distribution that uses $\{ f_{i} \}_{i=1}^{d}$ and a prior $P (x)$ over $\mathcal{X}$. 
This means that $Q$ is in the form $Q \in \mathcal{Q}$, where: 
\begin{align*}
\mathcal{Q}
:= 
\left\{ Q (x \mid \lambda) \in \Delta(\mathcal{X}): \exists \lambda \in \mathbb{R}^{d} : \;
Q (x \mid \lambda) \propto P (x) \exp \left( \sum_{i=1}^{d} \lambda_{i} f_{i} (x) \right) \right\}
\end{align*}

Depending on the definitions of the features $\{ f_{i} \}_{i=1}^{d}$, this can be extremely powerful and expressive. 
A choice of $\mathcal{Q}$ follows easily once $\{ f_{i} \}_{i=1}^{d}$ is chosen. 
Now the task at hand is modified to include $\mathcal{Q}$:

\textbf{Predict with a distribution $Q$ that minimizes the log loss to the data $\hat{P}_{n}$ (cross-entropy $\text{H} (\hat{P}_{n}, Q)$) within the appropriate exponential family model class $\mathcal{Q}$. }

This amounts to:
\begin{align*}
\boxed{\min_{Q \in \mathcal{Q}} \text{H} (\hat{P}_{n}, Q)}
\end{align*}

which is a very standard and universal way of prescribing learning. 
This precisely describes unsupervised learning; supervised learning corresponds to using conditional distributions $\text{Pr} (y \mid x)$ in this same formalism, with $P(x)$ known. 
Other learning scenarios map similarly on to this framework, only changing the space $\mathcal{X}$ and model class $\mathcal{Q}$. 

This exponential family formalism will be central to us for many reasons that will be discussed -- for one thing, the information projection $P_{\mathcal{A}}^{*}$ is in $\mathcal{Q}$. 
For now, it's enough to realize it as a model class that dictates the learning problem, in this prescription of learning. 

This formulation can be interpreted as minimizing the description length of the model on the data. 
The minimum description length principle (philosophically, "Occam's Razor") has been useful for learning for a long time, with roots ranging from likelihood maximization to Kolmogorov complexity and compressibility \cite{grunwald2007minimum}.

\subsubsection{Prescription II: minimizing "robust Bayes" log loss}

Another way to think about learning from limited observations is to accept that our observations about $\hat{P}_{n}$ don't uniquely determine $P$. 
So what is our optimization function?
The best we can do is an upper bound on loss, over distributions in $\mathcal{A}$ that satisfy our observations. 
This leads to another guiding justification for learning: 

\textbf{Predict with a distribution $Q$ that minimizes the log loss to the data-generating distribution $P$ (cross-entropy $\text{H} (P, Q)$), given $P \in \mathcal{A}$ (i.e., given our observations about $\hat{P}_{n}$ hold for $P$). }

This amounts to behaving as if the objective function is $\max_{P \in \mathcal{A}} \text{H} (P, Q)$, the upper bound on loss we have described. 
Therefore, the loss minimization problem faced by the learner becomes:
\begin{align*}
\boxed{
V_{\mathcal{A}} := \min_{Q \in \Delta(\mathcal{X})} \max_{P \in \mathcal{A}} \text{H} (P, Q)
}
\end{align*}

Clearly, our loss minimization problem is $\min_{Q \in \Delta(\mathcal{X})} \text{H} (\hat{P}_{n}, Q) \leq V_{\mathcal{A}}$, so $V_{\mathcal{A}}$ is a tight bound on our log loss. 
This is known as a "robust Bayes" approach \cite{GD04}. 
The optimal $Q$ here is not trivial, depending sensitively on the structure of $\mathcal{A}$. 

Information theory studies these ideas with a slightly different focus, relating similar robustness concepts in information and coding theory. 
What robust Bayes and statistical mechanics aim to optimize is the worst-case loss, which is deeply related to the information-theory notion of \emph{redundancy} of a communication channel. 
A milestone theorem of information theory says that this is equal to the capacity of the channel -- the "redundancy-capacity theorem" \cite{merhav1995strong, haussler1997general}.

\subsubsection{Prescription III: maximizing probability of the observations}

An intuitively appealing principle guiding modeling is to predict with the distribution that makes the data most likely. 

This is the principle behind maximum-likelihood modeling, but there the modeling restrictions are expressed in terms of a model class. 
In our situation, we are instead guided by the information $\mathcal{A}$ restricting the data. 
We can formulate the learning principle directly: 

\textbf{Predict with a distribution that maximizes the probability of the given observations, i.e. a distribution $\hat{P}_{n}$ maximizing $ \frac{1}{n} \log \text{Pr} \left( \hat{P}_{n} \in \mathcal{A} \right)$.}

We have seen that this amounts to solving 
\begin{align*}
\boxed{
\min_{Q \in \mathcal{A}} \text{D} (Q \Vert P)
}
\end{align*}
because of Sanov-type behavior (\Cref{sec:genprobcalc}). 
The KL divergence, which emerges from Boltzmann's calculation as intimately linked to probability, is the objective function for learning here.

\subsection{Equivalence of these prescriptions}
\label{sec:subequivalencelearning}

All these perspectives lead to exactly the same learning problem! 
This is the key observation of this section. 

We show this by showing that the commonly used log loss minimization with $\mathcal{Q}$ is exactly following probability maximization, then showing that robust Bayes and probability maximization each require maximizing entropy in the same manner.

\subsubsection{Probability maximization $\iff$ Log loss minimization with $\mathcal{Q}$}

The probability calculation-based learning problem is equivalent to the log loss minimization with the exponential family model class. 

Starting with the latter, the log loss minimization problem is:
\begin{align*}
\arg\min_{Q \in \mathcal{Q}} \text{H} (\hat{P}_{n} , Q) 
\end{align*}

It is clear because of the definition of divergence $\text{D} (\hat{P}_{n} \Vert Q) = \text{H} (\hat{P}_{n} , Q) - \text{H} (\hat{P}_{n})$ that: 
$ \displaystyle
\arg\min_{Q \in \mathcal{Q}} \text{H} (\hat{P}_{n} , Q) 
= \arg\min_{Q \in \mathcal{Q}} \text{D} (\hat{P}_{n} \Vert Q) 
$. 

Finally, we use the fact that 
\begin{align*}
\arg\min_{Q \in \mathcal{Q}} \text{D} (\hat{P}_{n} \Vert Q) 
= 
\arg\min_{Q \in \mathcal{A}} \text{D} (Q \Vert P)
\end{align*}
which we prove later in \Cref{sec:subappest}. 
The right-hand side $\displaystyle \arg\min_{Q \in \mathcal{A}} \text{D} (Q \Vert P)$ is exactly the probability calculation. 

This shows the equivalence of log loss minimization over $\mathcal{Q}$ and probability maximization.

\subsubsection{Probability maximization $\iff$ Maximum entropy}

The probability-maximization principle is equivalent to maximizing entropy under a uniform prior.  

As we have seen, a uniform distribution over $\mathcal{X}$ is a common and intuitive choice of prior (\Cref{sec:unifprior}). 
Writing this prior as $P_{0}$, the learning problem is 
\begin{align*}
\min_{P \in \mathcal{A}} \text{D} (P \Vert P_{0}) = \min_{P \in \mathcal{A}} \left[ \text{H} (P , P_{0}) - \text{H} (P) \right]
&\iff  \max_{P \in \mathcal{A}} \text{H} (P) 
\end{align*}
because $\text{H} (P , P_{0})$ is the same for all $P$. 
This key property of the uniform prior makes it the special, unique prior under which max-entropy is equivalent to the probability calculation.

\subsubsection{Robust Bayes $\iff$ Maximum entropy}
The robust Bayes prescription for learning is to solve 
\begin{align*}
V_{\mathcal{A}} := \min_{Q \in \Delta(\mathcal{X})} \max_{P \in \mathcal{A}} \text{H} (P, Q)
\end{align*}

Using the minimax theorem (e.g. that of \cite{S58}), we can swap the order of the min and max here, giving: 
\begin{align*}
V_{\mathcal{A}} = \max_{P \in \mathcal{A}} \min_{Q \in \Delta(\mathcal{X})} \text{H} (P, Q)
\end{align*}

Since the best possible distribution to describe $P$ is $P$ itself, we see that $V_{\mathcal{A}}$ is the maximum of the entropy: 
\begin{align*}
V_{\mathcal{A}} = \max_{P \in \mathcal{A}} \min_{Q \in \Delta(\mathcal{X})} \text{H} (P, Q) = \max_{P \in \mathcal{A}} \text{H} (P, P) = \max_{P \in \mathcal{A}} \text{H} (P)
\end{align*}

So optimizing over a tight upper bound on log loss is completely equivalent to the problem of maximizing entropy. 
This has long been well known \cite{DPDPL97}, and the max-ent problem's solutions -- the exponential family distributions -- are used everywhere.

\subsection{Interpretation: one learning principle to rule them all}

As we see, these perspectives are all equivalent! 
They amount to different faces of learning, each having a different significance: 
\begin{itemize}
\item 
The first perspective is the operational one taken by much of practical machine learning today. 
Minimizing log loss to the data, using a model class to regularize and attain generalization, is the dominant approach. 
An exponential family is a somewhat restrictive model class, but also quite general, as dictated by observations. 

\item
The robust Bayes perspective gives an interpretation of this as maximum entropy, and as being robust in various ways. 
It is flexible and generalizable -- it shows how two evidence sets $\mathcal{A}$ can be combined by intersection in the optimization problem. 

\item
The probability-calculation-based perspective shows how generalization occurs from the sample $\hat{P}_{n}$ to $P$. 
This is a crucial basis to the premise that we can learn from samples, 
and a useful way to interpret the impact of changes to the learning problem. 
\end{itemize}

In addition, the maximum entropy problem 
\begin{align*}
V_{\mathcal{A}} = \max_{P \in \mathcal{A}} \text{H} (P)
\end{align*}
is also equivalent to all these problems under a uniform prior -- so the many philosophical interpretations for max-entropy are of interest whenever one of these learning problems is solved. 
The max-entropy principle has been championed for learning since the early days of the modern field, notably by Jaynes \cite{jaynes1957information, jaynes1957information2, jaynes1968prior, jaynes1979concentration, jaynes1982rationale}. 
This

When the prior is not uniform, the maximum entropy principle generalizes to a "minimum discrimination information" (MDI) principle for some prior $P$ and some posterior $Q$, with the following learning problem: 
\begin{align*}
\boxed{\boxed{
\min_{Q \in \mathcal{A}} \text{D} (Q \Vert P)
}}
\end{align*} 
In a sense, this is the most basic and general perspective. 
It has been repeatedly proposed as the correct generalization of max-entropy \cite{kullback1959information}. 
We've seen in \Cref{sec:subgenprobcalc} that MDI is in accordance with making the "most likely" event happen from prior $P$ under $\mathcal{A}$, because of Sanov-like behavior. 

The MDI principle goes back a long way, as Gauss used it to derive the form of the normal distribution \cite{campbell1970equivalence}. 
The MDI principle has been connected to the prescriptions of learning mentioned earlier \cite{topsoe1993game}, again starting with the discrete case \cite{good1963maximum}. 
There is a line of well-established work robustly expressing this as a two-player zero-sum game as we have, bridging information theory and game theory \cite{harremoes2001maximum}. 
Our development builds on these concepts with exact probability calculations, to connect with Sanov's theorem.

\subsection{Axiomatic perspectives: why log loss?}

The axiomatic basis of entropy and cross-entropy is now a widely studied topic. 
Early work established the pattern, and is normally credited to Shannon \cite{shannon1948} in his landmark development of information theory, and to Khinchin \cite{khinchin1949mathematical} in statistical mechanics. 
These "Shannon-Khinchin axioms" were the foundation of significant pioneering work since then \cite{aczel1974shannon}. 

MDI-like perspectives have been apparent since this early work. 
Information theorists were very aware of it through its natural and unique role in Sanov-type concentration behavior \cite{cover1994processes}, providing the most likely evolution of a prior $P$ conditioned on evidence $\mathcal{A}$ \cite{van1981maximum}. 
On a different note, there were some powerful attempts to justify MDI and cross-entropy through a set of postulates, using logic similar to that for entropy \cite{shore1980axiomatic, skilling1988axioms, csiszar1991least}.

I think it's very instructive to sketch the main broad arguments that justify $\text{D} (Q \Vert P)$ as the only possible function to minimize to learn $Q$ from $P$ (and $\mathcal{A}$). 
(For simplicity, I'll take $\mathcal{X}$ to be discrete.) 
Some of the arguments are intuitive but not widely known.  

In all cases we describe below, our observations are given by $\mathcal{A}$, and so we're looking at the learning problem as solving 
\begin{align*}
\min_{Q \in \mathcal{A}} \text{F} (Q \Vert P)
\end{align*} 
for some divergence function $\text{F}$. 
We will see that this function must be the KL divergence under some natural conditions.

\subsubsection{Axioms: aggregating evidence}

One vital property of $\text{D} (Q \Vert P)$ is how it behaves recursively when we observe finer-grained outcomes. 
We can think of this as splitting two outcomes in a distribution $P = (P_1, P_2, P_3, \dots, P_d) \in \mathbb{R}^{d}$ to make $P' = (P^{a}_1, P^{b}_1, P_2, P_3, \dots, P_d) \in \mathbb{R}^{d+1}$, with $P^{a}_1 + P^{b}_1 = P_1$. 

Suppose the same thing is done to a distribution $Q = (Q_1, Q_2, \dots, Q_d) \in \mathbb{R}^{d}$ to make $Q' = (Q^{a}_1, Q^{b}_1, Q_2, \dots, Q_d) \in \mathbb{R}^{d+1}$. 
Then: 
\begin{align*}
\text{D} (Q \Vert P) = \text{D} (Q' \Vert P') + Q_1 \text{D} ( [Q^{a}_1, Q^{b}_1] \Vert [P^{a}_1, P^{b}_1] )
\end{align*} 

This property is extremely intuitive; it just means that adding the extra outcome to $\text{D} (Q' \Vert P')$ conveys information recursively according to how the outcomes are split, weighted according to the amount of probability involved. 
We would like this postulate to hold for any reasonable divergence function $\text{F}$. 

If we use this postulate repeatedly on $\text{F}$, defining $S_i = \sum_{j=1}^{i} Q_i$ and $R_i = \sum_{j=1}^{i} P_i$, 
\begin{align*}
\text{F} (Q \Vert P) = \sum_{i=2}^{n} S_i \text{F} \left( \left[ \frac{S_{i-1}}{S_{i}} , \frac{Q_{i}}{S_{i}} \right] \Vert \left[ \frac{R_{i-1}}{R_{i}} , \frac{P_{i}}{R_{i}} \right] \right)
\end{align*} 
which helpfully reduces the problem to distributions over two outcomes, and gives a lot of structure to any $\text{F}$ following the postulate above. 

It can be shown that any $\text{F}$ following this and a few more intuitive postulates must be the relative entropy $\text{D} (Q \Vert P)$. 
And due to the recursivity above, it is enough to check these other postulates for distributions over just a couple of outcomes. 

These other postulates on $\text{F}$ hold for any reasonable divergence measure. 
They are: 
\begin{itemize}
\item 
$\text{F}$ is zero when the arguments are the same (a distribution is always zero distance to itself). 
\item 
$\text{F}$ is regular, having well-behaved derivatives. 
\item 
$\text{F}$ is independent of the way in which the outcomes are labeled. 
\end{itemize}

Any function following these is a constant multiple of the relative entropy $\text{D} (Q \Vert P)$ (\cite{hobson1969new, kannappan1973characterization}, see also \cite{faddeev1956concept, renyi1961measures}). 

This is emblematic of Shannon's \cite{shannon1948} axiomatic approach, where most of the power of the result is derived from the assumed additivity of $\text{F}$ under combination of independent sources of evidence. 
It provides a rigorous reason why $\text{F}$ is tensorized over the $n$ examples, and why each component takes the $q \log \frac{q}{p}$ form (for more, see \cite{skilling1988axioms}).

\subsubsection{Axioms: invariance principles}

Another approach to seeing the necessity of $\text{D} (Q \Vert P)$ relies on invariance principles \cite{johnson1979axiomatic}. 

The argument goes that any reasonable learning rule must produce the same results: (a) if the parameter space is transformed; (b) when two independent systems are considered jointly or separately; and (c) "whether one treats an independent subset of system states in terms of a separate conditional density or in terms of the full system density" \cite{shore1980axiomatic}. 
These appealing axioms, about the invariance of the learning procedure using $\text{F}$, imply that it must be the relative entropy $\text{D} (Q \Vert P)$ \cite{shore1980axiomatic}. 

This is remarkable, because there's no explicit assumption about any functional form in the axioms. 
However, there are sometimes reasons to doubt their applicability (for instance, we might expect parameter transformations to materially affect learning).  
In such situations, another axiomatic justification, like the one above based on aggregating evidence, might be preferred.

\subsubsection{Axioms: maximizing probability}

It turns out that cross-entropy is the only loss with the probability-based perspective we've discussed in \Cref{sec:subequivlearning}, which makes it the only loss capable of counting probabilities consistently \cite{tikochinsky1984alternative}. 
This is an extremely powerful result, which can be understood based on our discussions so far. 

We have justified $\text{D} (Q \Vert P)$ and $\text{H} (Q, P)$ from first principles by computing probabilities. 
Observe that this alternate probability-based perspective can't be rewritten in any different form, because the probability calculations we use (including the general ones) are exact identities. 
Hence, no per-sample loss on observations $Q$ can count probabilities under $P$ consistently other than $\text{D} (Q \Vert P)$.

\subsection{Discussion: a united learning paradigm}

As we have seen, this discussion -- and the max-entropy problem -- intimately involves \emph{exponential families}. 
For predictive modeling with loss minimization, they matter for several reasons related to our previous equivalent views of learning with an exponential family $\mathcal{Q}$. 

We've seen that exponential families give the commonly used log loss minimization a special significance. 
Minimizing log loss on observed data within $\mathcal{Q}$ maximizes the probability of the data given $\mathcal{A}$. 
This probability-calculation viewpoint is very fundamental, giving machine learning a rigorous and unique grounding due to realities of statistical mechanics: the behavior of concentration under expected-value observations. 

Additionally, another equivalent view is the robust Bayes optimization problem, where we minimize the worst-case loss over all possible data distributions satisfying the observations: 
\begin{align*}
\arg\min_{Q \in \mathcal{Q}} \text{H} (\hat{P}_{n}, Q) 
= \arg\min_{Q \in \Delta(\mathcal{X})} \max_{P \in \mathcal{A}} \text{H} (P, Q) 
\end{align*}
This is another view that comports nicely with worst-case techniques used to robustify modern learning, featuring heavily in works from distributionally robust optimization \cite{duchi2021learning, sagawa2019distributionally}, adversarial learning \cite{goodfellow2014explaining}, and other robust learning scenarios \cite{arjovsky2019invariant, madry2018towards, mohri2019agnostic}. 
The exponential family solutions are maximally robust, in a way that is easy to appreciate through $\mathcal{A}$.

What we're describing is true at all regimes, with no approximations, for any $n$ and $d$. 
So it can be used to study overfitting and generalization, high-dimensional regimes, and everything in between. 
The difference is only in the specified constraint values $\alpha$. 

Incredibly, all of this can be derived from a streamlined starting point: defining a set of real-valued feature functions, considering observations to be their averages over the sample, and calculating probabilities. 
The consequences are extremely powerful -- complementary perspectives on a probability-based framework for learning. 

The leanness of the assumptions matters philosophically as well, as this framework is very general and applies to any data distribution, featurization, and learning problem.
There is no wrong choice of features -- we are discussing results which hold always, with no approximation, and can universally be applied to any modeling scenario, with different interpretations in each case.

There are no substantive restrictions on the space, allowing it to be very structured, beyond data like words. 
These ideas have had long-standing success on complex spaces $\mathcal{X}$, like the trajectory space of evolutions over time (the "Schrödinger bridge" problem \cite{leonard2014survey, chen2021stochastic, shi2024diffusion}), time series spectra \cite{burg1975maximum}, and more \cite{landau1987maximum}.

\section{Information projections and exponential families}

We've seen all sorts of justifications for the information projection of $P$ on $\mathcal{A}$, $P_{\mathcal{A}}^{*}$:  
\begin{align*}
\boxed{
P_{\mathcal{A}}^{*} := \arg\min_{Q \in \mathcal{A}} \text{D} (Q \Vert P)
}
\end{align*} 
It is time to explore it more explicitly. 
To get the explicit form, we can solve the original constrained optimization problem: 
$$
\begin{aligned}
\min_{Q \in \Delta (\mathcal{X})} \text{D} (Q \Vert P) 
\quad \text{s.t.} & \quad \mathbb{E}_{Q} [ f_i(x) ] = \alpha_i \quad \text{for all } i = 1, \dots, d.
\end{aligned}
$$

Using the technique of Lagrange multipliers to enforce the constraints, the distribution that solves this is given by:
\begin{align*}
P_{\mathcal{A}}^{*} (x) \propto P (x) \exp \left( \sum_{i=1}^{d} \lambda_i^* f_i(x) \right)
\end{align*}

where $\left\{ \lambda_i^* \right\}_{i=1}^{d}$ are the Lagrange multipliers that ensure $P_{\mathcal{A}}^{*} \in \mathcal{A}$.\footnote{We'll assume these exist, so that the problem is well-posed. (See \cite{follmer2011stochastic}, Thm. 3.24 for the calculation.)} 

The normalization is expressed through the \textit{log-partition function}
$ \displaystyle
A (\lambda) = \log \mathbb{E}_{P} \left[ \exp \left( \sum_{i=1}^{d} \lambda_i f_i(x) \right) \right]
$, 
so that $P_{\mathcal{A}}^{*} (x) = P (x) \exp \left( \sum_{i=1}^{d} \lambda_i^* f_i(x) - A (\lambda^{*}) \right)$.

$P_{\mathcal{A}}^{*}$ is unique in all non-degenerate cases, and has a convenient closed form parametrized by $\lambda \in \mathbb{R}^{d}$. 
The family of such distributions with varying $\lambda$ is called the exponential family associated to features $f (x) \in \mathbb{R}^{d}$ and prior $P$:\footnote{If no prior is given, a uniform prior is typically assumed.} 
\begin{align*}
\mathcal{Q}
:= 
\left\{ Q (x \mid \lambda) \in \Delta(\mathcal{X}): \exists \lambda \in \mathbb{R}^{d} : \;
Q (x \mid \lambda) \propto P (x) \exp \left( \sum_{i=1}^{d} \lambda_{i} f_{i} (x) \right) \right\}
\end{align*}

From these definitions, observe that $P_{\mathcal{A}}^{*} \in \mathcal{A} \cap \mathcal{Q}$.\footnote{We'll ignore issues of $\mathcal{Q}$ being an open/closed set here; otherwise all the results we show are true with $\mathcal{Q}$ being replaced by its closure.}

\section{Statistical mechanics: energy and loss minimization}

We arrive at the task we've been building towards -- laying out the properties of this learning task using a rigorous statistical mechanics viewpoint, drawing a web of connections between statistical physics, information theory, and learning. 

This starts with the equivalent perspectives of \Cref{sec:seclearning}, which shed some light on the quantities $\text{H} (P), \text{H} (P, Q), \text{D} (P \Vert Q)$, traditionally thought of as being from information-theory. 

From a log loss minimization perspective, $\text{H} (P, Q)$ is the loss being minimized. 
Meanwhile, the quantities 
\begin{align*}
\text{H} (P) = \min_{S \in \Delta (\mathcal{X})} \text{H} (P, S) 
\qquad
\text{D} (P \Vert Q) = \text{H} (P, Q) - \min_{S} \text{H} (P, S)
\end{align*} 
are nicely interpretable. 
$\text{H} (P)$ is the Bayes loss -- the loss suffered by any predictor on this problem. 
The divergence $\text{D} (P \Vert Q)$ is evidently interpretable as the regret with respect to the inherent (Bayes) loss in describing the data $P$. 

As this illustrates, the definitions of statistical mechanics have broad quantitative meaning, but their names are sometimes not illuminating in our context. 
Now we carry over those physics-influenced definitions to our loss-minimization setting.

\subsection{Internal energy and the log-partition function}

Define the \textbf{internal energy} that the exponential family associates with any microstate $x$ and feature $i$ as 
$$ U_i^{\lambda} (x) := - \lambda_i f_i (x)$$ 
This is the energy of the system in state $x$ associated with the $i$-th feature, which adds to make the total energy $ U^{\lambda} (x) := \sum_{i=1}^{d} U_i^{\lambda} (x) $, 
so that $P_{\lambda} (x) \propto \exp \left( - \sum_{i=1}^{d} U_i^{\lambda} (x) \right) = \exp \left( - U^{\lambda} (x) \right)$. 
Taken over distributions, we have 
$$ U^{\lambda} (P) := - \sum_{i=1}^{d} \lambda_i \mathbb{E}_{x \sim P} [f_i (x)] = \mathbb{E}_{x \sim P} \left[ \sum_{i=1}^{d} U_i^{\lambda} (x) \right] $$ 

At equilibrium (when the expected moments match the empirical moments), the internal energy is $- \sum_{i=1}^{d} \lambda_i \alpha_i$ where $\alpha_i$ are the supplied constraints. 

The log-partition function is:
\begin{align*}
A ( \lambda ) &:= \min_{Q \in \Delta(\mathcal{X})} \;\max_{P \in \Delta(\mathcal{X})} \left[ \text{H} (P, Q) - U^{\lambda} (P) \right]
\end{align*}

This is a good way to \textit{define} the log-partition function from a machine learning perspective. 
It illuminates a central zero-sum game being played between the min-player and the max-player, which is a mutually unconstrained Lagrangian version of the robust Bayes setup introduced earlier. 

If we use the minimax theorem \cite{S58} to swap the order of the min and max, 
\begin{align*}
A ( \lambda ) &:= \max_{P \in \Delta(\mathcal{X})} \;\min_{Q \in \Delta(\mathcal{X})} \left[ \text{H} (P, Q) - U^{\lambda} (P) \right] \\
&= \max_{P \in \Delta(\mathcal{X})} \left[ \text{H} (P) + \sum_{i=1}^{d} \lambda_{i} \mathbb{E}_{x \sim P} [ f_{i} (x) ] \right]
\end{align*}

Since $A$ is a maximum over linear functions of $\lambda$, it is convex in $\lambda$.

\subsection{Free energy}

The \textbf{free energy} of any distribution $P$ is the difference between its internal energy and its entropy: 
\begin{align*}
F^{\lambda} (P) := U^{\lambda} (P) - \text{H}(P) = - \sum_i \lambda_i \mathbb{E}_{x \sim P} [f_i (x)] - \text{H}(P)
\end{align*}

Therefore, for any exponential family distribution $P_{\lambda}$, 
\begin{align*}
F^{\lambda} (P_{\lambda}) = - A (\lambda)
\end{align*}

In the language of constrained optimization, the free energy is the Lagrangian of the constrained maximization of entropy -- indeed, $A (\lambda)$ is the Fenchel dual of the negative entropy function. 
It has a more physically intuitive interpretation too in the study of thermodynamics, involving concepts of work and heat, which can be connected to our statistical view later.

\subsection{Entropy}

In general, the entropy is a minimum of linear functions by definition -- $\text{H} (P) = \min_{S \in \Delta (\mathcal{X})} \text{H} (P, S)$ where $\text{H} (P, S) = \mathbb{E}_{x \sim P} \text{H} (x, S)$. 
Therefore, $H (P)$ is concave in $P$. 

The entropy of the exponential family distribution $P_{\lambda}$ is 
\begin{align*}
\text{H} (P_{\lambda}) 
&:= U^{\lambda} (P_{\lambda}) - F^{\lambda} (P_{\lambda}) 
= \sum_{i=1}^{d} - \lambda_i \mathbb{E}_{x \sim P_{\lambda}} \left[ f_i(x) \right] + A (\lambda)
= - \sum_{i=1}^{d} \lambda_i \alpha_i + A (\lambda)
\end{align*}

This is convex in $\lambda$; when it is being maximized over a convex constraint set $\mathcal{A}$, this means that the entropy-maximizing $P_{\lambda^{*}} = P_{\mathcal{A}}^{*}$ lies at the boundary of the set $\mathcal{A}$. 

% If two such exponential family distributions $P_1 , P_2$ are mixed, the maximum entropy distribution will still be by far the most probable. 
% So the averaged distribution $S = \frac{1}{2} (P_{1} + P_{2})$ has maximal entropy, i.e. $H (S) \geq \frac{1}{2} ( H (P_{1}) + H (P_{2}) )$. 

\subsection{Loss and internal energy}

For any distribution $P$ whatsoever, and any parameters $\lambda$, the loss $\mathbb{E}_{x \sim P} \left[ - \log (P_{\lambda} (x)) \right] = \text{H} (P, P_{\lambda})$ is: 
\begin{align}
\label{eq:defofcrossent}
\text{H} (P, P_{\lambda})
=
A (\lambda) - \sum_i \lambda_i \mathbb{E}_{x \sim P} [f_i (x)]
= U^{\lambda} (P) - F^{\lambda} (P_{\lambda})
\end{align}

In other equivalent words, for any $\lambda$: 
\begin{align*}
\text{H} (x, P_{\lambda}) = - \sum_{i=1}^{d} \lambda_{i} f_{i} (x) + A (\lambda) = U^{\lambda} (x) - F^{\lambda} (P_{\lambda})
\end{align*}

(To get some intuition on this, if the loss is low, $U^{\lambda} (x) \approx F^{\lambda} (P_{\lambda})$, which is highly negative. 
So the energy is generally negative over the data.) 

Therefore, for any two arbitrary distributions $P, Q$: 
\begin{align*}
\text{H} (P, P_{\lambda}) - \text{H} (Q, P_{\lambda})
= U^{\lambda} (P) - U^{\lambda} (Q)
\end{align*}

Therefore, $U^{\lambda} (P) = \text{H} (P, P_{\lambda}) + K$ for some constant $K$. \footnote{The constant must result in $U^{0} (P) = 0$, i.e. $K = - \text{H} (P, P_{0})$ where $P_{0}$ is the uniform distribution.}

This shows that the internal energy $U^{\lambda} (P)$ of data $P$ corresponds to the loss incurred by predicting with $P_{\lambda}$ on $P$. 
It is low when $P_{\lambda}$ is a good approximation of the data, and higher otherwise.

\subsection{Regret and free energy}

Subtracting $\text{H}(P)$ from both sides of the equation \eqref{eq:defofcrossent}, we get that for any distribution $P$, 
\begin{align*}
\text{D} \left( P \Vert P_{\lambda} \right) = F^{\lambda} (P) - F^{\lambda} (P_{\lambda})
\end{align*}

For us, the free energy therefore corresponds to the model's regret. 
It's the excess loss, over what we would incur if we knew $P$. 
If the model does as well as $P$ at describing the data, the regret/free energy will be low, even if $P$ is noisy. 

In the language of duality in statistical physics, this is the dual interpretation (minimum free energy at fixed "temperature" $\lambda$) to the usual primal variational characterization (maximum entropy at fixed internal energy).

\subsection{Mean-field approximation: Bogoliubov's inequality}

Exponential families have another convenient property for learning. 
Suppose we are looking to approximate an exponential family distribution with another "\textit{variational}" model distribution, possibly in a different exponential family, with different features and parameters $\psi$. 

As part of an exponential family, we can write this $P_{\psi} (x) = \exp \left( - U^{\psi} (x) - A (\psi) \right)$. 
For any such variational $P_{\psi}$, we have 
\begin{align*}
0 &\leq \text{D} (P_{\psi} \Vert P_{\lambda}) \\
&= \mathbb{E}_{x \sim P_{\psi} } \left[ - U^{\psi} (x) - A (\psi) + U^{\lambda} (x) + A (\lambda) \right] \\
&= (U^{\lambda} (P_{\psi}) - U^{\psi} (P_{\psi})) + (F^{\psi} (P_{\psi}) - F^{\lambda} (P_{\lambda}))
\end{align*}

Thus, if $\psi$ is chosen to have $U^{\lambda} (P_{\psi}) = U^{\psi} (P_{\psi})$, then the free energy $F^{\psi} (P_{\psi}) \geq F^{\lambda} (P_{\lambda})$. 
In this case, the free energy of $P_{\psi}$ can be used as a tight bound on the free energy of the unknown distribution $P_{\lambda}$, as long as $U^{\lambda}$ and $U^{\psi}$ are the same under the variational distribution $P_{\psi}$. 

Similarly, a variational lower bound $F^{\psi} (P_{\psi}) \leq F^{\lambda} (P_{\lambda})$ holds as long as $U^{\lambda}$ and $U^{\psi}$ are the same under the target distribution $P_{\lambda}$. 
\begin{align*}
0 &\geq - \text{D} (P_{\lambda} \Vert P_{\psi}) \\
&= - \mathbb{E}_{x \sim P_{\lambda} } \left[ - U^{\lambda} (x) - A (\lambda) + U^{\psi} (x) + A (\psi) \right] \\
&= (U^{\lambda} (P_{\lambda}) - U^{\psi} (P_{\lambda})) + (F^{\psi} (P_{\psi}) - F^{\lambda} (P_{\lambda}))
\end{align*}

Note that conditions on $U (P)$ are often more amenable to computation than $H (P)$ or $F (P)$, since expectation values of observables can be typically computed easily from finite samples. 

This is a basic principle from statistical mechanics underlying mean-field variational inference. 
It is often called Bogoliubov's inequality.

\subsection{On these definitions}

We have chosen these definitions carefully to ensure these properties, and to correspond to a general constrained loss minimization problem. 
These are not always the standard definitions we see when information theory and statistics content mentions statistical physics. 
There are also other ways of proceeding; for instance, the divergence $\text{D} ( \cdot \Vert \cdot )$ can be defined as the objective of the game \cite{GD04}. 
The definitions we present here are chosen to result in definitions of all the basic statistical mechanics and thermodynamics laws.

\section{Exponential families and their properties}
\label{sec:expfamex}

Everything we have described so far leads to the key role of the distribution $P_{\mathcal{A}}^{*}$ over $\mathcal{X}$. 
It is the target distribution according to all the perspectives of learning we mentioned earlier: log loss minimization over exponential family $\mathcal{Q}$, robust Bayes under observations $\mathcal{A}$, and probability maximization of the observed data. 

\Cref{sec:boltzmanncalc} has shown that the basic quantities of information theory, like (cross/relative) entropy, emerge from first principles when calculating probabilities of fluctuations in sampling from a discrete distribution. 
The calculation, performed first by Boltzmann, shows that high-entropy configurations are exponentially more likely than other configurations. 

This happens to an overwhelming degree in the macroscopically sized samples of atoms we observe in everyday life: our observations are almost deterministic, even of highly disordered systems like gaseous and liquid matter. 

From a modeling perspective, exponential family distributions are extremely useful, as we have described in \Cref{sec:seclearning}. 
It's worth describing them a little more comprehensively, with a view to setting up the statistical mechanics interpretation of these learning problems, and unlock some more of their power for the prediction problems fundamental to machine learning.

Here I collect properties of exponential families that are useful for learning, all derived from the above definitions.

\subsection{Sufficient statistics and factorizations}

A major motivating principle behind exponential families is that they are exactly the distributions with sufficient statistics whose dimension does not grow with the amount of data, i.e. a parametric model. 
This is known as the Pitman-Koopman-Darmois theorem on sufficient statistics \cite{fisher1922mathematical, koopman1936distributions, Darmois1935, pitman1936sufficient}, and is a foundational result we will not discuss further here. 

A similarly central place is taken by the factorization theorem for probability densities characterizing sufficient statistics, also studied in depth around that time \cite{neyman1935suunteorema} and shown to be a very general perspective on sufficiency \cite{halmos1949application}. 
The general idea of the proof is interesting and fundamental
\footnote{Here is a short proof for the discrete case $\mathcal{X} = \{ X_1, \dots, X_n \}$. 
In this case, the factorization theorem says that any statistic $T(X)$ is sufficient if and only if the probability mass function (PMF) of $X$ can be factored as $f_n (X|\theta) = u(X)v(T(X), \theta)$ for some functions $u$ and $v$. 
We prove both directions of the implication in turn. 

First, assume that the PMF of $X$ can be factored as above. Then:
$ \displaystyle P(T = t) = \sum_{x:T(x) = t} P(X = x) = v(t, \theta) \sum_{x:T(x)=t} u(x) $. 
The conditional distribution of $X$ given $T$ is 
$ \displaystyle
P(X = x | T = t) = \frac{P(X = x, T = t)}{P(T = t)} = \frac{u(x)}{\sum_{x:T(x)=t} u(x)}
$. 
Since this expression does not depend on $\theta$, $T(X)$ is a sufficient statistic.

Conversely, if we start by assuming $T(X)$ is sufficient, we can factor 
$
P(X = x | \theta) = P(X = x | T = t)  P(T = t | \theta)
$. 
Let $ u(x) = P(X = x | T = t) $ and $ v(t, \theta) = P(T = t | \theta) $, giving the factorization $ P(X = x | \theta) = u(x)v(T(x), \theta) $. 
This proves the converse direction and the result.}, 
shown to hold for exponential families. 

Exponential families are vital for graph-based inference as well. 
For any graph structure (CRF), the distributions that factorize over its cliques in the graph structure. Furthermore  that factorize like this are exactly the ones that obey the Markov property with respect to the graph. 
This is often called the Hammersley-Clifford theorem, or the Gibbs-Markov theorem \cite{georgii2011gibbs}. 
Though the result is extremely unique and general, its proof is essentially straightforward, and is given in \cite{Pollard04, besag1974spatial}. 
It also means that if we are satisfied with exponential family conditional distributions on each edge of the graph (we have settled on some edge-wise features), we can efficiently model joint distributions over the graph. 
Useful practical graphs like trees have sparse connectivity, admitting practically informative factorizations and efficient algorithms \cite{chow1968approximating}. 

%see https://originalstatic.aminer.cn/misc/billboard/aml/Hammersley-Clifford%20Theorem.pdf , 

\subsection{Data-generating "robustness"}

Exponential families are remarkably easy to compare to each other with the divergence $\text{D} ( \cdot \Vert \cdot )$.  
For any distributions $P_{\alpha}, P_{\beta}$ from the same exponential family $\mathcal{Q}$, and any distribution $Q$, we have 

\begin{align*}
\text{D} (Q \Vert P_{\beta}) - \text{D} (Q \Vert P_{\alpha}) 
&= \mathbb{E}_{x \sim Q} \left[ \log \left( P_{\alpha} (x) \right) - \log \left( P_{\beta} (x) \right) \right] \\
&= \mathbb{E}_{x \sim Q} \left[ \left( \sum_{i=1}^{d} \alpha_i f_i(x) - A (\alpha) \right) - \left( \sum_{i=1}^{d} \beta_i f_i(x) - A (\beta) \right) \right] \\
&= \sum_{i=1}^{d} (\alpha_i - \beta_i ) \mathbb{E}_{x \sim Q} \left[ f_i(x) \right] - A (\alpha) + A (\beta)
\end{align*}

This key equation has a few important consequences when $Q \in \mathcal{A}$. 
Suppose $Q, P_{\alpha} \in \mathcal{A}$, i.e. the data follow the same moment constraints as one of the distributions. 
Then, 
\begin{align}
\label{eq:expfamrobustness}
\text{D} (Q \Vert P_{\beta}) - \text{D} (Q \Vert P_{\alpha}) 
&= \sum_{i=1}^{d} (\alpha_i - \beta_i ) \mathbb{E}_{x \sim Q} \left[ f_i(x) \right] - A (\alpha) + A (\beta) \nonumber \\
&= \sum_{i=1}^{d} (\alpha_i - \beta_i ) \mathbb{E}_{x \sim P_{\alpha}} \left[ f_i(x) \right] - A (\alpha) + A (\beta) \nonumber \\
&= \mathbb{E}_{x \sim P_{\alpha}} \left[ \log \left( \frac{P_{\alpha} (x)}{P_{\beta} (x)} \right) \right] 
= \text{D} (P_{\alpha} \Vert P_{\beta})
\end{align}

This is called "robustness" of exponential families \cite{grunwald2007minimum}: 
the relative performance of two coding schemes $P_{\alpha}, P_{\beta}$ is the same when measured by any $Q \in \mathcal{A}$. 
In our situation, it means that if $\hat{P}_{n}$ denotes the observed data distribution and $P_{\mathcal{A}}^{*}$ the max-entropy distribution under the observed features, 
then for any $P_{\lambda} \in \mathcal{Q}$, 
\begin{align*}
\text{D} (\hat{P}_{n} \Vert P_{\lambda}) - \text{D} (\hat{P}_{n} \Vert P_{\mathcal{A}}^{*}) 
&= \text{D} (P_{\mathcal{A}}^{*} \Vert P_{\lambda})
\end{align*}

In other words, for the task of predicting the observed $\hat{P}_{n}$, the relative performance of any model $P_{\lambda}$ to the best $P_{\mathcal{A}}^{*}$ is just $\text{D} (P_{\mathcal{A}}^{*} \Vert P_{\lambda})$. 

\textbf{The relative prediction loss of any exponential family distribution $P_{\lambda}$ to the best $P_{\mathcal{A}}^{*}$ is $\text{D} (P_{\mathcal{A}}^{*} \Vert P_{\lambda})$, regardless of any other details of the data $\hat{P}_{n}$. }

This has been shown with "performance" being measured by regret. 
Note that  
\begin{align*}
\text{D} (Q \Vert P_{\beta}) - \text{D} (Q \Vert P_{\alpha})
= 
\text{H} (Q , P_{\beta}) - \text{H} (Q , P_{\alpha})
\end{align*}
so all these statements are true for relative loss as well.

\subsection{Approximation and estimation error}
\label{sec:subappest}

% Substituting any two distributions $Q_1, Q_2 \in \mathcal{A}$ into $(*)$ above: 
% \begin{align*}
% \text{D} (Q_1 \Vert P_{\gamma}) - \text{D} (Q_1 \Vert P_{\lambda}) &- \text{D} (Q_2 \Vert P_{\gamma}) + \text{D} (Q_2 \Vert P_{\lambda}) \\
% &= \mathbb{E}_{x \sim Q_1} \left[ \log \left( \frac{P_{\lambda} (x)}{P_{\gamma} (x)} \right) \right] 
% - \mathbb{E}_{x \sim Q_2} \left[ \log \left( \frac{P_{\lambda} (x)}{P_{\gamma} (x)} \right) \right] \\
% &= \sum_{i=1}^{d} (\lambda_i - \gamma_i ) \left( \mathbb{E}_{x \sim Q_1} \left[ f_i(x) \right] - \mathbb{E}_{x \sim Q_2} \left[ f_i(x) \right] \right) 
% = 0
% \end{align*}

%Setting $Q_1 = P_{\gamma} = P_{\mathcal{A}}^{*}$ for any $P_{\mathcal{A}}^{*} \in \mathcal{A} \cap \mathcal{Q}$, 

Setting $P_{\alpha}$ to be $P_{\mathcal{A}}^{*} \in \mathcal{A} \cap \mathcal{Q}$ in the equation \eqref{eq:expfamrobustness} above, we get a very useful result about this max-entropy distribution $P_{\mathcal{A}}^{*}$: the divergence satisfies a Pythagorean theorem for any $P$ meeting the moment constraints ($P \in \mathcal{A}$), and any $P_{\lambda} \in \mathcal{Q}$. 
\begin{align*}
\forall P \in \mathcal{A} , P_{\lambda} \in \mathcal{Q} : \qquad
\underbrace{\text{D} (P \Vert P_{\lambda}) }_{\text{regret}} = \underbrace{\text{D} (P_{\mathcal{A}}^{*} \Vert P_{\lambda})}_{\text{estimation error}} + \underbrace{\text{D} (P \Vert P_{\mathcal{A}}^{*})}_{\text{approximation error}}
\end{align*}

This is a decomposition of the relative loss (the regret) into estimation and approximation errors. 

\begin{itemize}
\item
The approximation error is lowered by considering more expressive architectures. 
\item
The estimation error is lowered by considering more data. 
\end{itemize}

This means that if all we know about the data is encapsulated in $\mathcal{A}$, it is a good idea to minimize over the parametric family $\mathcal{Q}$ (under the geometry induced by $\text{D}$). 
There are some specific consequences to the Pythagorean equality above. 

In the Pythagorean equality, we clearly see that both the approximation and estimation errors are $\geq 0$. 
Applying this understanding gives us two inequalities, which hold for any $\mathcal{A}$ and associated $\lambda$. 

First, the overall regret exceeds the estimation error: 
\begin{align*}
\text{D} (P \Vert P_{\lambda}) \geq \text{D} (P_{\mathcal{A}}^{*} \Vert P_{\lambda}) \quad \forall P \in \mathcal{A} , P_{\lambda} \in \mathcal{Q}
\end{align*}

This can be readily interpreted -- 
for any exponential family model $P_{\lambda}$, the actual data is harder to encode than the max-ent distribution. 

On the other hand, the overall regret also evidently exceeds the approximation error: 
\begin{align*}
\text{D} (P \Vert P_{\lambda}) \geq \text{D} (P \Vert P_{\mathcal{A}}^{*}) \quad \forall P \in \mathcal{A} , P_{\lambda} \in \mathcal{Q}
\end{align*}

Since the data $\hat{P}_{n} \in \mathcal{A}$ by definition, this applies to them: 
$\text{D} (\hat{P}_{n} \Vert P_{\lambda}) \geq \text{D} (\hat{P}_{n} \Vert P_{\mathcal{A}}^{*}) \quad \forall \lambda$. 
As $P_{\mathcal{A}}^{*}$ is in the exponential family $\mathcal{Q}$, this means that 
\begin{align*}
P_{\mathcal{A}}^{*} = \arg\min_{P_{\lambda} \in \mathcal{Q}} \; \text{D} (\hat{P}_{n} \Vert P_{\lambda})
\end{align*}
which shows that $P_{\mathcal{A}}^{*}$ minimizes the log loss (cross entropy) to the data over $\mathcal{Q}$, as discussed earlier (\Cref{sec:subequivalencelearning}).

\subsection{The estimation error and deviance}

The estimation error relates to the concept of \textbf{deviance}, which uses the divergence $\text{D} (\cdot \Vert \cdot)$ to relate a non-equilibrium probability distribution $P_{\lambda}$ to the equilibrium distribution $P_{\lambda^{*}}$. 
We can evaluate the ratio of the distributions at any given observed feature representation $f(x) \in \mathbb{R}^{d}$: 
\begin{align*}
\log \left( \frac{P_{\lambda} (f(x))}{P_{\lambda^{*}} (f(x))} \right)
= \sum_{i=1}^{d} (\lambda_i - \lambda_i^* ) f_i(x) + A (\lambda^{*}) - A (\lambda)
\end{align*}
At the actual observation $f(x) = \alpha$, 
\begin{align*}
\log \left( \frac{P_{\lambda} (\alpha)}{P_{\lambda^{*}} (\alpha)} \right) 
&= \sum_{i=1}^{d} (\lambda_i - \lambda_i^* ) \alpha_i + A (\lambda^{*}) - A (\lambda) 
= - \text{D} (P_{\lambda^{*}} \Vert P_{\lambda}) 
\end{align*}
which exactly shows how suboptimal parameter settings will deviate around the optimum in modeling the observations.

\subsection{The approximation error and entropy}
\label{sec:subapproxerror}

How well does the information projection $P_{\mathcal{A}}^{*}$ approximate the data $P$?

For any data meeting the constraints, i.e. $P \in \mathcal{A}$, and any $\lambda$: 
\begin{align*}
\text{D} (P \Vert P_{\mathcal{A}}^{*}) 
&= \text{H} (P, P_{\lambda}) - \text{H} (P_{\mathcal{A}}^{*} , P_{\lambda}) + \text{H} (P_{\mathcal{A}}^{*}) - \text{H} (P) 
\end{align*}

In particular this is true for $\lambda = 0$, in which case $P_{\lambda} = P_{\lambda = 0} = P_{0}$ (the uniform distribution over the data), 
so $\text{H} (P, P_{0}) = \text{H} (P_{\mathcal{A}}^{*} , P_{0})$, and this reduces to 
$$
\text{D} (P \Vert P_{\mathcal{A}}^{*}) 
= \text{H} (P_{\mathcal{A}}^{*}) - \text{H} (P) 
$$ 

Therefore, the data will be well approximated if they have roughly maximal entropy under the constraints. 
Tying together these concepts, the max-entropy problem has a variational characterization: $\text{H} (P_{\lambda^{*}}) - \text{H} (P) = \text{D} (P \Vert P_{\lambda^{*}})$ for all $P$ matching the moment constraints. 
This extends to any moment constraints, so we could also say for any $\lambda$ that 
$
\text{D} (P \Vert P_{\lambda}) = \text{H} (P_{\lambda}) - \text{H} (P)
$, 
for all $P$ having the same feature moments as $P_{\lambda}$.

\subsection{Evaluating exponential family models}

Using this in the regret decomposition above, 
\begin{align*}
\underbrace{\text{D} (P \Vert P_{\lambda}) }_{\text{regret}} &= \text{D} (P_{\mathcal{A}}^{*} \Vert P_{\lambda}) + \text{H} (P_{\mathcal{A}}^{*}) - \text{H} (P) \\
&= \text{H} (P_{\mathcal{A}}^{*} , P_{\lambda}) - \text{H} (P) 
\end{align*}

Adding $\text{H} (P)$ to both sides gives an interesting result: 
\begin{align*}
\forall P \in \mathcal{A} : \qquad \qquad \text{H} (P , P_{\lambda}) &= \text{H} (P_{\mathcal{A}}^{*} , P_{\lambda})
\end{align*}

The interpretation here is unambiguous: 
\textbf{for evaluating the loss using the exponential family $\mathcal{Q}$, we can pretend the data follows $P_{\mathcal{A}}^{*}$. }

\subsection{Using data to approximate the exponential family}

We can flip the roles of $P$ and $P_{\mathcal{A}}^{*}$ in the above question about divergence: how well does the data $P$ approximate $P_{\mathcal{A}}^{*}$?

It turns out that: 
\begin{align*}
- \text{D} (P_{\mathcal{A}}^{*} \Vert P) 
\geq \frac{1}{n} \log \text{Pr} ( \hat{P}_{n} \in \mathcal{A} ) 
\geq - \text{H} (P_{\mathcal{A}}^{*} , P )
\end{align*}

So if $\hat{P}_{n}$ is consistent with the observations $\mathcal{A}$ and $\text{Pr} ( \hat{P}_{n} \in \mathcal{A} )$ is high, then $\text{D} (P_{\mathcal{A}}^{*} \Vert P)$ is quite low - the data $P$ is a good approximation of samples generated with $P_{\mathcal{A}}^{*}$. 

To show the lower bound here, we use the Sanov-type probability identity and the fact that $\mu_{\mathcal{A}} \in \mathcal{A}$: 
\begin{align*}
\frac{1}{n} \log \text{Pr} ( \hat{P}_{n} \in \mathcal{A} ) 
&= - \text{D} (P_{\mathcal{A}}^{*} \Vert P ) - \frac{1}{n} \text{D} ( \mu_{\mathcal{A}} \Vert P_{\mathcal{A}}^{*n} ) \\
&= - \text{H} (P_{\mathcal{A}}^{*} , P ) + \text{H} ( P_{\mathcal{A}}^{*} ) - \frac{1}{n} \text{H} ( \mu_{\mathcal{A}} , P_{\mathcal{A}}^{*n} ) + \frac{1}{n} \text{H} ( \mu_{\mathcal{A}} ) \\
&= - \text{H} (P_{\mathcal{A}}^{*} , P ) + \frac{1}{n} \text{H} ( P_{\mathcal{A}}^{*n} ) - \frac{1}{n} \text{H} ( P_{\mathcal{A}}^{*n} ) + \frac{1}{n} \text{H} ( \mu_{\mathcal{A}} ) 
\geq - \text{H} (P_{\mathcal{A}}^{*} , P )
\end{align*}

% <!-- $$ \text{D} (P_1 \Vert Q_1) + \text{D} (P_2 \Vert Q_2) - \text{D} (P_1 \Vert Q_2) - \text{D} (P_2 \Vert Q_1) = \lambda ( \mathbb{E}_{x \sim P_1} [f] - \mathbb{E}_{x \sim P_2} [f] ) $$ -->

\subsection{The log-partition function and higher moments}

A well-known result \cite{chowdhury2023bregman} connects the cumulant-generating function of the features under an exponential family distribution $P_{\lambda} \propto \exp \left( \sum_{i=1}^{d} \lambda_i f_i(x) \right)$ to the log-partition function $A (\lambda)$. 
\begin{align*}
\log \mathbb{E}_{x \sim P_{\lambda}} \left[ \exp \left( \sum_{i=1}^{d} \theta_i f_i(x) \right) \right] 
&= A (\lambda + \theta) - A (\lambda)
\end{align*}
This also implies a cumulant-generating function for the centered features, i.e. those with mean zero, which is in the form of a Bregman divergence $\text{B}_{F} (P, Q) := F (P) - F (Q) - (P-Q)^\top \nabla F (Q)$: 
\begin{align*}
\log \mathbb{E}_{x \sim P_{\lambda}} \left[ \exp \left( \sum_{i=1}^{d} \theta_i \left( f_i(x) - \mathbb{E}_{x \sim P_{\lambda}} [f_i (x)] \right) \right) \right] 
&= A (\lambda + \theta) - A (\lambda) - \theta^\top \nabla A (\lambda)
= \text{B}_{A} (\lambda + \theta, \lambda)
\end{align*}

We can now discuss the special relationship between higher moments and the Fisher information. 
This starts with differentiating $A$. 
Writing $Z := \exp (A)$: 
\begin{align*}
\frac{\partial A (\lambda)}{\partial \lambda_{i}} 
&= \frac{1}{Z} \frac{\partial Z (\lambda)}{\partial \lambda_{i}} 
= \frac{1}{Z} \mathbb{E} \left[ f_i (x) \exp \left( \sum_{i=1}^{d} \lambda_i f_i (x) \right) \right]
= \mathbb{E}_{x \sim P_{\lambda}} \left[ f_i(x) \right]
\end{align*}

Differentiating $A$ again, we get the Fisher information $I (\lambda)$: 
\begin{align*}
\frac{\partial^2 A (\lambda)}{\partial \lambda_{i} \partial \lambda_{j}} 
&= \frac{\partial \mathbb{E}_{x \sim P_{\lambda}} \left[ f_i (x) \right]}{\partial \lambda_{j}} 
= \frac{\partial \mathbb{E}_{x \sim P_{\lambda}} \left[ f_j (x) \right]}{\partial \lambda_{i}} \\
&= \mathbb{E}_{x \sim P_{\lambda}} \left[ f_i (x) f_j (x) \right] - \mathbb{E}_{x \sim P_{\lambda}} \left[ f_i (x) \right] \mathbb{E}_{x \sim P_{\lambda}} \left[ f_j (x) \right] \\
&= \text{cov}_{x \sim P_{\lambda}} \left[ f_i (x), f_j (x) \right] \\
&= \mathbb{E}_{x \sim P_{\lambda}} \left[ \left( f_i (x) - \mathbb{E}_{x \sim P_{\lambda}} \left[ f_i (x) \right] \right) \left( f_j (x) - \mathbb{E}_{x \sim P_{\lambda}} \left[ f_j (x) \right] \right) \right] \\
&= \mathbb{E}_{x \sim P_{\lambda}} \left[ \left( \frac{\partial [\log P_{\lambda} (x)]}{\partial \lambda_{i}} \right) \left( \frac{\partial [\log P_{\lambda} (x)]}{\partial \lambda_{j}} \right) \right] 
= - \mathbb{E}_{x \sim P_{\lambda}} \left[ \frac{\partial^2 [\log P_{\lambda} (x)]}{\partial \lambda_{i} \partial \lambda_{j}} \right] \\
&:= [I (\lambda)]_{i, j}
\end{align*}

\subsection{Fluctuations}

In \Cref{sec:fluctuationsgaussian}, we discussed how to view the probability distribution over less-visited states. 
Now that we have an understanding of this distribution, we can analyze fluctuations \cite{touchette2009large} relative to the equilibrium distribution over microstates $\mathcal{X}$, which we already know we can treat as $P_{\mathcal{A}}^{*}$. 
Write this as $P_{\lambda^{*}}$.

Statistical physicists often approximate this by assuming the probability distribution near the optimum to be roughly Gaussian, with covariance $[I (\lambda^{*})]^{-1}$. 
This idea -- often called Laplace's approximation -- has been repeatedly used in both statistical physics and AI/ML \cite{mackay1992bayesian, williams1998bayesian}. 
Extending this idea for any observable, fluctuations in concentration can be seen as fluctuations in free energy, which can be used to express well-known concentration behavior in learning \cite{maurer2012thermodynamics}.

% There are many such physics-based quantitative connections to learning that emerge from this. 
For any feature, the learner observes a fixed average feature value, and considers different parameter settings $\lambda$. 
The sensitivity of the observation to parameter changes can be identified with the "heat capacity" of a particular feature:
\begin{align*}
\frac{\partial [ \mathbb{E}_{x \sim P_{\lambda}} \left[ f_i(x) \right] ]}{\partial T_i } 
= - \lambda_i^2 \frac{\partial [ \mathbb{E}_{x \sim P_{\lambda}} \left[ f_i(x) \right] ]}{\partial \lambda_i } 
= - \lambda_i^2 \text{var}_{x \sim P_{\lambda}} \left[ f_i (x) \right]
\leq 0
\end{align*}
In our language, decreasing the temperature (increasing the "coolness" $\lambda_i$) tends to raise $\mathbb{E}_{x \sim P_{\lambda}} \left[ f_i(x) \right]$. 
There is intuition for this in statistical physics, where the heat capacity is identified with $\text{var}_{x \sim P_{\lambda}} \left[ f_i (x) \right]$, going back to \cite{einstein1904allgemeinen, peliti2017einstein}.

\section{Statistical mechanics as a bridge to physics}

Statistical physics has a deep relationship with the learning framework we have laid out. 
Though the starting motivations of the two are different, they cover the same topics, with a common statistical mechanics formalism discussed in \Cref{sec:seclearning}. 

\cite{jaynes1957information} said of statistical mechanics, "it is possible to maintain a sharp distinction between its physical and statistical aspects. The former consists only of the correct enumeration of the states of a system and their properties." 
In other words, the physics of the problem only show through in how $\mathcal{X}, P, \{ f_i (x) \}_{i=1}^{d}$ are specified. 
Instead, we can view these as related to loss minimization without physics aims, letting us unlock much of the power in learning scenarios. 

We've seen in \Cref{sec:seclearning} many complementary ways of viewing the learning problem. 
Though there is a deep unity between the ideas of statistical mechanics and our setting, some of the differences with statistical mechanics are in the interpretations of a common technical core. 

The statistical physics problem is motivated by microstate counting, i.e. probability maximization.
In contrast, the learning problem is motivated practically by the log loss minimization, and also has principled foundations as a minimax robust Bayes loss problem. 
We've seen in \Cref{sec:seclearning} that they are all equivalent -- once this is set up, the same unified stat-mech formalism governs behavior. 

Statistical physics differentiates between "extensive" variables, which grow in proportion to the scaling of the system (like mass, volume, entropy), and "intensive" variables, which are independent of the scaling of the system. 
In the view of statistical mechanics, "extensive" variables are replaced by observables and "intensive" ones by parameters -- the broad intuition is that intensive quantities are the $\left\{ f_i (x) \right\}_{i=1}^{d}$ model $\lambda$ values, while extensive quantities are functions of the observed data $\left\{ f_i (x) \right\}_{i=1}^{d}$. 
(Extensivity has a quantitatively precise meaning in thermodynamics, which is beyond scope here.)

AI/ML is applied to scenarios large and small, much more so than statistical physics. 
Accordingly, the approach we are describing is very general -- everything applies to general $n$ and $d$, and the framework can consider all manner of feature functions, and a variety of common and exotic regularization schemes. 
This level of powerful variety is often absent from other applications of statistical mechanics in observable physics, although present to an increasing degree in information theory and statistics. 

Notably, it can be simpler to look at things in general terms. 
Statistical physics has different statistical ensembles motivating the probability calculations -- the microcanonical, canonical, and grand canonical ensembles \cite{tolman1979principles} are instances of just one general constrained optimization stat-mech framework, which serves different purposes depending on interpretation.

\subsection{Statistical mechanics interpretations and privileged constraints}
\label{sec:subprivilegedconstraints}

One example is the choice of prior. 
As we've discussed in \Cref{sec:unifprior}, the fundamental link between entropy and multiplicity can be derived when there was underlying physics to motivate a uniform prior (Liouville's theorem in Hamiltonian mechanics, in that case). 
Our analogous "physics-like" basic principle is the empirical sampling process that leads to the training set. 
This is typically assumed to be i.i.d. with respect to the test distribution. 
But deviations like train-test distribution shifts could easily necessitate a different non-uniform prior. 
The exact probabilistic developments here are a foundation for more explorations into generalization across distribution shifts. 

Then there is the differing interpretation of particular constraints in the optimization. 
As described in \Cref{sec:fluctuationsgaussian}, energy is often a "privileged" constraint in physics contexts. 
In applications of statistical mechanics to physics, the energy constraint is treated specially \cite{landaulifshitz} -- it is the principal observable of interest through the Hamiltonian \cite{pathria2017statistical}. 
This difference of interpretation has significant consequences both in our approach and in statistical physics. 

Let's sketch the general viewpoint on this. 
The energy is a privileged constraint associated with an observable energy for each macrostate, which we call a feature $f_1$. 
$\mathcal{A}$ is just a $(d-1)$-dimensional space defined by observing the average $f_1$. 
A distribution in this exponential family is then given by $P_{\lambda} (x) \propto \exp \left( \lambda_1 f_1(x) \right)$. 
The Lagrange multiplier $\lambda_1$ is the "inverse temperature" or "coolness" associated with $f_1$, i.e. $T_1 := 1/\lambda_1$ is the temperature associated with $f_1$ \cite{baez2024entropy}. 
By convention, the internal energy associated with $f_1$ is $U_1 = \mathbb{E}_{x \sim P_{\lambda}} [f_1 (x)] = \mathbb{E}_{x \sim P_{\lambda}} [f_1 (x)] = - \frac{\partial }{\partial \lambda_1} [A (\lambda)]$ where $A$ is the log-partition function. 

There's a small difference in that the energy does not depend on a value of $\lambda_1$, only $f_1$. 
Instead, the free energy associated with $f_1$ under an exponential family $P_{\lambda}$ is $ = - \frac{1}{\lambda_1} [A (\lambda)]$, so the factor of $\lambda_1$ gets absorbed into the free energy. 
This is just a matter of convention, and doesn't materially affect our results. 

For historical reasons, there are many redundant quantities and potentials in statistical physics. 
For instance, considering different observation variables as measured and/or fixed leads to a huge variety of free energies. 
The Helmholtz, Gibbs, and other free energies are just derived by privileging the volume constraint (fixing its dual variable pressure), or the energy (fixing its dual, temperature), or the particle number of a species (fixing its dual, chemical potential of the species). 
All these fall under a similar unified formalism from our point of view -- addressing all the constraints, potentially interacting on the microstate (individual data example) level, applying to overlapping sets of microstates. 
% We can isolate generalization issues that occur here in AI with rare subsets of data (corresponding to chemical species). 

This work addresses static learning equilibria under uniform random sampling -- the statistical mechanics covered here has been applied to statistical physics in thermostatics \cite{pathria2017statistical}. 
When this scenario is perturbed, we get thermodynamics, and the most famous consequences of statistical mechanics. 
This is left to explore elsewhere.

\subsection{Unity through duality}

Statistical mechanics can written as constrained optimization, with the constraints supplied by observations. 
It has been exceptionally successful in describing various observations in physics, because it is both precise and general. 
Duality is a key optimization concept here that generalizes past physical situations to learning problems. 

In learning, all the constrained optimization problems we look at can be solved by introducing Lagrange multipliers for the constraints, which we consider to be parameters. 
Parameters are dual to observed variables -- each one measures the tightness of the constraint associated with that observation. 
A setting with maximally loose constraints is $\lambda = 0$, but other settings have a lower energy. 

Energies describe the Lagrangian, the objective of the unconstrained problem that is equivalent to the constrained one. 
The energy of a solution shows how much it strains against the constraints. 
Because of the way Lagrange duality works, energies add, even for constraints that interact with each other in possibly complex ways. 
This has allowed learning algorithms to freely manipulate the loss function through additive regularization terms. 

Early modern statistical mechanics developed the concept of "generalized forces" \cite{khinchin1949mathematical}. 
Generalized forces are the dual variables of the constraints -- in other words, they are the Lagrange parameters. 
They represent sensitivity of the system's energy (prediction performance) to other parameters \cite{kardar2007statistical, gao2019generalized}, and have been recognized under different names in various fields which use constrained optimization. 
For instance, they correspond to "shadow prices" in economics, parameters in computer science, and temperature or chemical potentials in physics. 

From a loss minimization perspective, cross entropy is arguably a more fundamental quantity than entropy or relative entropy. 
The cross entropy to the true data is the quantity being directly optimized in the learning formulations we've developed, corresponding to learning with minimum description length \cite{grunwald2007minimum, B17, hinton1993keeping}. 

Of course, in learning situations we are interested in other losses beyond cross-entropy. 
Some special cases have been solved in convenient closed form \cite{BF15, BF16, chow1968approximating, nguyen2017inverse, lynn2024exact}, and the technical framework exists for a comprehensive duality-based understanding \cite{topsoe1979information, GD04}.

\subsection{Philosophies of modeling}

These learning formulations represent a philosophical shift away from a very common way of thinking about data and modeling. 

The model is often thought of as an object whose parametric form is provided, which makes arbitrary assumptions on the parametric form of the solution. 
If a solution is not describable in these parameters, the model may be arbitrarily bad at figuring this out -- this is often called model misspecification. 
A model class is defined, and a focus for theory is how expressive this model class is. 
Practitioners do not have to worry about misspecification, and instead focus on learning: finding a good model in the class. 
All this is made possible by a range of parametric forms, typically used for statistics/ML convenience. 
The exponential family $\mathcal{Q}$ is sometimes thought of in these terms, as a parametric form used for convenience in various Bayesian calculations.

The stance suggested by the robust Bayes perspective of \Cref{sec:subequivlearning} is very different. 
The modeling, though it does ultimately involve $d$ parameters, does not make parametric assumptions about the data. 
Once features are chosen for a dataset, there is always a best $\lambda^*$ which satisfies the feature constraints at their observed data-derived values. 
So rather than thinking of the model's form and parameters as chosen based on a set of assumptions, the modeling assumptions can be stated implicitly, by defining observation functions (features) and setting up the robust Bayes minimax problem \cite{GD04}, which leaves the modeling distribution unconstrained. 
In many specific situations, this is a more natural and less arbitrary intuition for modeling. 

This is a very different way of thinking about data than the traditional parametric approach, where we assume a fixed number of parameters and try to estimate them using an assumed functional form. 
Here what is assumed is a set of observations -- we are given feature functions -- and not the model form. 

This reasoning ultimately privileges a particular distribution - the maximum-entropy distribution - over all others, as it is by far the most likely to have generated the data. 
For evaluating the performance, we can effectively treat the data as if it came from this distribution, and calculate deviations accordingly. 
To do this, we only need to specify our observations, not anything about the data-generating process. 

In other words, our choice to featurize and observe the data in a particular way is the only assumption we make about the data -- this \textbf{determines} the distribution of the data within strict bounds, exactly as our observations of atoms are constrained to certain known bulk properties. 

Another side of this is an interesting perspective on model misspecification. 
If a data distribution does not follow an exponential family model, this can always be addressed by adding more features (observation constraints) to the model, which increases performance and lowers the entropy of the learned distribution. 
As many features as necessary can be added. 
The learned distribution always still maximizes the probability of the observed data. 
To the extent that a feature really influences the solution, i.e. lowers the entropy of the distribution $P_{\mathcal{A}^{*}}$, it will lower $\text{H} (P_{\mathcal{A}^{*}}) - \text{H} (\hat{P}_{n}) = \text{D} (\hat{P}_{n} \Vert P_{\mathcal{A}^{*}})$, and bring down the approximation error accordingly, as we described in \Cref{sec:subapproxerror}. 
And even if the learning model is not predictive, it is still an information projection onto some observed $\mathcal{A}$, with all the interpretations involved in that (e.g. Sec. \ref{sec:seclearning}). 

Exploring this connection is a matter of ongoing work, interesting to us from the AI perspective because of the correspondence to feature learning. 
Boosting \cite{FS97, SF12} is a notable success of this approach in incremental feature learning, and many others apply information-theoretic principles to other representation learning settings \cite{globerson2003sufficient, B17, hjelm2019learning, kawaguchi2023does}. 
Feature learning through loss minimization has enjoyed amazing success in deep learning as well, with a huge variety of multipurpose embeddings providing solutions to an array of problems. 
Such developments in learning continue to broaden the applicability and versatility of the statistical mechanics framework for analyzing observed distributions.

% \chapter{Thermodynamics}
%\input{SMD_thermodynamics}
% \input{SMD_generalized_thermodynamics}

\newpage
\bibliographystyle{alpha}
\bibliography{main}

\newcommand{\etalchar}[1]{$^{#1}$}
\begin{thebibliography}{HFLM{\etalchar{+}}19}

\bibitem[ABGLP19]{arjovsky2019invariant}
Martin Arjovsky, L{\'e}on Bottou, Ishaan Gulrajani, and David Lopez-Paz.
\newblock Invariant risk minimization.
\newblock {\em arXiv preprint arXiv:1907.02893}, 2019.

\bibitem[AFN74]{aczel1974shannon}
J{\'a}nos Acz{\'e}l, Bruno Forte, and Che~Tat Ng.
\newblock Why the shannon and hartley entropies are ‘natural’.
\newblock {\em Advances in applied probability}, 6(1):131--146, 1974.

\bibitem[Bae24]{baez2024entropy}
John~C. Baez.
\newblock What is entropy?
\newblock {\em arXiv preprint arXiv:2409.09232}, 2024.

\bibitem[Bal17]{B17}
Akshay Balsubramani.
\newblock Optimal binary autoencoding with pairwise correlations.
\newblock In {\em Proceedings of the International Conference on Learning Representations (ICLR)}, 2017.
\newblock arXiv preprint arXiv:1611.02268.

\bibitem[Bal20]{balsubramani2020sharp}
Akshay Balsubramani.
\newblock Sharp finite-sample concentration of independent variables.
\newblock {\em arXiv preprint arXiv:2008.13293}, 2020.

\bibitem[Bes74]{besag1974spatial}
Julian Besag.
\newblock Spatial interaction and the statistical analysis of lattice systems.
\newblock {\em Journal of the Royal Statistical Society: Series B (Methodological)}, 36(2):192--225, 1974.

\bibitem[BF15]{BF15}
Akshay Balsubramani and Yoav Freund.
\newblock Optimally combining classifiers using unlabeled data.
\newblock In {\em Conference on Learning Theory (COLT)}, 2015.

\bibitem[BF16]{BF16}
Akshay Balsubramani and Yoav~S Freund.
\newblock Optimal binary classifier aggregation for general losses.
\newblock In {\em Advances in Neural Information Processing Systems}, pages 5032--5039, 2016.

\bibitem[BKM17]{blei2017variational}
David~M Blei, Alp Kucukelbir, and Jon~D McAuliffe.
\newblock Variational inference: A review for statisticians.
\newblock {\em Journal of the American statistical Association}, 112(518):859--877, 2017.

\bibitem[BKP{\etalchar{+}}20]{bahri2020statistical}
Yasaman Bahri, Jonathan Kadmon, Jeffrey Pennington, Sam~S Schoenholz, Jascha Sohl-Dickstein, and Surya Ganguli.
\newblock Statistical mechanics of deep learning.
\newblock {\em Annual Review of Condensed Matter Physics}, 11(1):501--528, 2020.

\bibitem[Bol77]{boltzmann1877beziehung}
Ludwig Boltzmann.
\newblock {\em {\"U}ber die Beziehung zwischen dem zweiten Hauptsatze des mechanischen W{\"a}rmetheorie und der Wahrscheinlichkeitsrechnung, respective den S{\"a}tzen {\"u}ber das W{\"a}rmegleichgewicht}, volume~76.
\newblock Kk Hof-und Staatsdruckerei, 1877.

\bibitem[Bur75]{burg1975maximum}
John~Parker Burg.
\newblock Maximum entropy spectral analysis.
\newblock {\em PhD thesis, Stanford University}, 1975.

\bibitem[Cam70]{campbell1970equivalence}
LL~Campbell.
\newblock Equivalence of gauss's principle and minimum discrimination information estimation of probabilities.
\newblock {\em The Annals of Mathematical Statistics}, 41(3):1011--1015, 1970.

\bibitem[CGP21]{chen2021stochastic}
Yongxin Chen, Tryphon~T Georgiou, and Michele Pavon.
\newblock Stochastic control liaisons: Richard sinkhorn meets gaspard monge on a schrodinger bridge.
\newblock {\em Siam Review}, 63(2):249--313, 2021.

\bibitem[CL68]{chow1968approximating}
CKCN Chow and Cong Liu.
\newblock Approximating discrete probability distributions with dependence trees.
\newblock {\em IEEE transactions on Information Theory}, 14(3):462--467, 1968.

\bibitem[Cov94]{cover1994processes}
Thomas~M Cover.
\newblock Which processes satisfy the second law.
\newblock {\em Physical origins of time asymmetry}, pages 98--107, 1994.

\bibitem[Csi84]{csiszar1984sanov}
Imre Csisz{\'a}r.
\newblock Sanov property, generalized $ i $-projection and a conditional limit theorem.
\newblock {\em The Annals of Probability}, 12(3):768--793, 1984.

\bibitem[Csi91]{csiszar1991least}
Imre Csiszar.
\newblock Why least squares and maximum entropy? an axiomatic approach to inference for linear inverse problems.
\newblock {\em The annals of statistics}, pages 2032--2066, 1991.

\bibitem[CSMG23]{chowdhury2023bregman}
Sayak~Ray Chowdhury, Patrick Saux, Odalric Maillard, and Aditya Gopalan.
\newblock Bregman deviations of generic exponential families.
\newblock In {\em The Thirty Sixth Annual Conference on Learning Theory}, pages 394--449. PMLR, 2023.

\bibitem[CT06]{CT06}
Thomas~M Cover and Joy~A Thomas.
\newblock {\em Elements of Information Theory}.
\newblock Wiley Series in Telecommunications and Signal Processing, 2006.

\bibitem[Dar35]{Darmois1935}
Georges Darmois.
\newblock Sur les lois de probabilité a estimation exhaustive.
\newblock {\em C. R. Acad. Sci. Paris}, 260:1265--1266, 1935.

\bibitem[DN21]{duchi2021learning}
John~C Duchi and Hongseok Namkoong.
\newblock Learning models with uniform performance via distributionally robust optimization.
\newblock {\em The Annals of Statistics}, 49(3):1378--1406, 2021.

\bibitem[DPDPL97]{DPDPL97}
Stephen Della~Pietra, Vincent Della~Pietra, and John Lafferty.
\newblock Inducing features of random fields.
\newblock {\em IEEE transactions on pattern analysis and machine intelligence}, 19(4):380--393, 1997.

\bibitem[DZ09]{dembozeitouni2009large}
Amir Dembo and Ofer Zeitouni.
\newblock {\em Large deviations techniques and applications}.
\newblock Springer, 2009.

\bibitem[Ein04]{einstein1904allgemeinen}
Albert Einstein.
\newblock Zur allgemeinen molekularen theorie der w{\"a}rme.
\newblock {\em Annalen der Physik}, 319(7):354--362, 1904.

\bibitem[Ein10]{einstein1910theorie}
Albert Einstein.
\newblock Theorie der opaleszenz von homogenen fl{\"u}ssigkeiten und fl{\"u}ssigkeitsgemischen in der n{\"a}he des kritischen zustandes.
\newblock {\em Annalen der Physik}, 338(16):1275--1298, 1910.

\bibitem[Ell99]{ellis1999theory}
Richard~S Ellis.
\newblock The theory of large deviations: from boltzmann’s 1877 calculation to equilibrium macrostates in 2d turbulence.
\newblock {\em Physica D: Nonlinear Phenomena}, 133(1-4):106--136, 1999.

\bibitem[Fad56]{faddeev1956concept}
Dmitrii~Konstantinovich Faddeev.
\newblock On the concept of entropy of a finite probabilistic scheme.
\newblock {\em Uspekhi Matematicheskikh Nauk}, 11(1):227--231, 1956.

\bibitem[Fis22]{fisher1922mathematical}
Ronald~A Fisher.
\newblock On the mathematical foundations of theoretical statistics.
\newblock {\em Philosophical transactions of the Royal Society of London. Series A, containing papers of a mathematical or physical character}, 222(594-604):309--368, 1922.

\bibitem[FS97]{FS97}
Yoav Freund and Robert~E. Schapire.
\newblock A decision-theoretic generalization of on-line learning and an application to boosting.
\newblock {\em J. Comput. Syst. Sci.}, 55(1):119--139, 1997.

\bibitem[FS11]{follmer2011stochastic}
Hans F{\"o}llmer and Alexander Schied.
\newblock {\em Stochastic finance: an introduction in discrete time}.
\newblock Walter de Gruyter, 2011.

\bibitem[GD04]{GD04}
Peter~D Gr{\"u}nwald and A~Philip Dawid.
\newblock Game theory, maximum entropy, minimum discrepancy and robust bayesian decision theory.
\newblock {\em The Annals of Statistics}, 32(4):1367--1433, 2004.

\bibitem[Geo11]{georgii2011gibbs}
Hans-Otto Georgii.
\newblock {\em Gibbs measures and phase transitions}.
\newblock Walter de Gruyter GmbH \& Co. KG, Berlin, 2011.

\bibitem[GGR19]{gao2019generalized}
Xiang Gao, Emilio Gallicchio, and Adrian~E Roitberg.
\newblock The generalized boltzmann distribution is the only distribution in which the gibbs-shannon entropy equals the thermodynamic entropy.
\newblock {\em The Journal of chemical physics}, 151(3), 2019.

\bibitem[Gib02]{gibbs1902elementary}
Josiah~Willard Gibbs.
\newblock {\em Elementary principles in statistical mechanics: developed with especial reference to the rational foundations of thermodynamics}.
\newblock C. Scribner's sons, 1902.

\bibitem[Goo63]{good1963maximum}
Irving~J Good.
\newblock Maximum entropy for hypothesis formulation, especially for multidimensional contingency tables.
\newblock {\em The Annals of Mathematical Statistics}, pages 911--934, 1963.

\bibitem[Gr{\"u}07]{grunwald2007minimum}
Peter~D Gr{\"u}nwald.
\newblock {\em The minimum description length principle}.
\newblock MIT press, 2007.

\bibitem[GSS14]{goodfellow2014explaining}
Ian~J Goodfellow, Jonathon Shlens, and Christian Szegedy.
\newblock Explaining and harnessing adversarial examples.
\newblock {\em arXiv preprint arXiv:1412.6572}, 2014.

\bibitem[GT03]{globerson2003sufficient}
Amir Globerson and Naftali Tishby.
\newblock Sufficient dimensionality reduction.
\newblock {\em Journal of Machine Learning Research}, 3(Mar):1307--1331, 2003.

\bibitem[HAK{\etalchar{+}}22]{huembeli2022physics}
Patrick Huembeli, Juan~Miguel Arrazola, Nathan Killoran, Masoud Mohseni, and Peter Wittek.
\newblock The physics of energy-based models.
\newblock {\em Quantum Machine Intelligence}, 4(1):1, 2022.

\bibitem[Hau97]{haussler1997general}
David Haussler.
\newblock A general minimax result for relative entropy.
\newblock {\em IEEE Transactions on Information Theory}, 43(4):1276--1280, 1997.

\bibitem[HFLM{\etalchar{+}}19]{hjelm2019learning}
R~Devon Hjelm, Alex Fedorov, Samuel Lavoie-Marchildon, Karan Grewal, Phil Bachman, Adam Trischler, and Yoshua Bengio.
\newblock Learning deep representations by mutual information estimation and maximization.
\newblock In {\em Proceedings of the International Conference on Learning Representations (ICLR)}, 2019.
\newblock arXiv preprint arXiv:1808.06670.

\bibitem[Hob69]{hobson1969new}
Arthur Hobson.
\newblock A new theorem of information theory.
\newblock {\em Journal of Statistical Physics}, 1:383--391, 1969.

\bibitem[HS49]{halmos1949application}
Paul~R Halmos and Leonard~J Savage.
\newblock Application of the radon-nikodym theorem to the theory of sufficient statistics.
\newblock {\em The Annals of Mathematical Statistics}, 20(2):225--241, 1949.

\bibitem[HT01]{harremoes2001maximum}
Peter Harremo{\"e}s and Flemming Tops{\o}e.
\newblock Maximum entropy fundamentals.
\newblock {\em Entropy}, 3(3):191--226, 2001.

\bibitem[HVC93]{hinton1993keeping}
Geoffrey~E Hinton and Drew Van~Camp.
\newblock Keeping neural networks simple by minimizing the description length of the weights.
\newblock In {\em Proceedings of the sixth annual conference on Computational learning theory}, pages 5--13, 1993.

\bibitem[Jay57a]{jaynes1957information}
Edwin~T Jaynes.
\newblock Information theory and statistical mechanics.
\newblock {\em Physical review}, 106(4):620, 1957.

\bibitem[Jay57b]{jaynes1957information2}
Edwin~T Jaynes.
\newblock Information theory and statistical mechanics. ii.
\newblock {\em Physical review}, 108(2):171, 1957.

\bibitem[Jay68]{jaynes1968prior}
Edwin~T Jaynes.
\newblock Prior probabilities.
\newblock {\em IEEE Transactions on systems science and cybernetics}, 4(3):227--241, 1968.

\bibitem[Jay79]{jaynes1979concentration}
Edwin~T Jaynes.
\newblock Concentration of distributions at entropy maxima.
\newblock {\em ET Jaynes: Papers on probability, statistics and statistical physics}, page 315, 1979.

\bibitem[Jay82]{jaynes1982rationale}
Edwin~T Jaynes.
\newblock On the rationale of maximum-entropy methods.
\newblock {\em Proceedings of the IEEE}, 70(9):939--952, 1982.

\bibitem[Jay86]{jaynes1986monkeys}
Edwin~Thompson Jaynes.
\newblock Monkeys, kangaroos and n.
\newblock {\em Maximum-Entropy and Bayesian Methods in Applied Statistics}, 26, 1986.

\bibitem[JGJS99]{jordan1999introduction}
Michael~I Jordan, Zoubin Ghahramani, Tommi~S Jaakkola, and Lawrence~K Saul.
\newblock An introduction to variational methods for graphical models.
\newblock {\em Machine learning}, 37:183--233, 1999.

\bibitem[Joh79]{johnson1979axiomatic}
R~Johnson.
\newblock Axiomatic characterization of the directed divergences and their linear combinations.
\newblock {\em IEEE Transactions on Information Theory}, 25(6):709--716, 1979.

\bibitem[Kar07]{kardar2007statistical}
Mehran Kardar.
\newblock {\em Statistical physics of particles}.
\newblock Cambridge University Press, 2007.

\bibitem[KDJH23]{kawaguchi2023does}
Kenji Kawaguchi, Zhun Deng, Xu~Ji, and Jiaoyang Huang.
\newblock How does information bottleneck help deep learning?
\newblock In {\em International Conference on Machine Learning}, pages 16049--16096. PMLR, 2023.

\bibitem[Khi49]{khinchin1949mathematical}
Aleksandr~Iakovlevich Khinchin.
\newblock {\em Mathematical foundations of statistical mechanics}.
\newblock Courier Corporation, 1949.

\bibitem[Koo36]{koopman1936distributions}
Bernard~Osgood Koopman.
\newblock On distributions admitting a sufficient statistic.
\newblock {\em Transactions of the American Mathematical society}, 39(3):399--409, 1936.

\bibitem[KR73]{kannappan1973characterization}
Palaniappan Kannappan and Pushpa~N Rathie.
\newblock On a characterization of directed divergence.
\newblock {\em Inf. Control.}, 22(2):163--171, 1973.

\bibitem[Kul59]{kullback1959information}
Solomon Kullback.
\newblock Information theory and statistics.
\newblock {\em Wiley, New York}, 1959.

\bibitem[KW{\etalchar{+}}19]{kingma2019introduction}
Diederik~P Kingma, Max Welling, et~al.
\newblock An introduction to variational autoencoders.
\newblock {\em Foundations and Trends{\textregistered} in Machine Learning}, 12(4):307--392, 2019.

\bibitem[Lan87]{landau1987maximum}
Henry~J Landau.
\newblock Maximum entropy and the moment problem.
\newblock {\em Bulletin of the American Mathematical Society}, 16(1):47--77, 1987.

\bibitem[LCH{\etalchar{+}}06]{lecun2006tutorial}
Yann LeCun, Sumit Chopra, Raia Hadsell, M~Ranzato, and Fujie Huang.
\newblock A tutorial on energy-based learning.
\newblock {\em Predicting structured data}, 1(0), 2006.

\bibitem[L{\'e}o14]{leonard2014survey}
Christian L{\'e}onard.
\newblock A survey of the schr{\"o}dinger problem and some of its connections with optimal transport.
\newblock {\em Discrete \& Continuous Dynamical Systems-A}, 34(4):1533--1574, 2014.

\bibitem[LL69]{landaulifshitz}
Lev~Davidovich Landau and Evgenii~Mikhailovich Lifshitz.
\newblock {\em Statistical Physics: Volume 5 Part 1}, volume~5.
\newblock Pergamon Press, 1969.

\bibitem[LN02]{leonard2002extension}
Christian L{\'e}onard and Jamal Najim.
\newblock An extension of sanov's theorem: application to the gibbs conditioning principle.
\newblock {\em Bernoulli}, pages 721--743, 2002.

\bibitem[LW19]{lamont2019correspondence}
Colin~H LaMont and Paul~A Wiggins.
\newblock Correspondence between thermodynamics and inference.
\newblock {\em Physical Review E}, 99(5):052140, 2019.

\bibitem[LYP{\etalchar{+}}24]{lynn2024exact}
Christopher~W Lynn, Qiwei Yu, Rich Pang, Stephanie~E Palmer, and William Bialek.
\newblock Exact minimax entropy models of large-scale neuronal activity.
\newblock {\em arXiv preprint arXiv:2402.00007}, 2024.

\bibitem[Mac92]{mackay1992bayesian}
David~JC MacKay.
\newblock Bayesian interpolation.
\newblock {\em Neural computation}, 4(3):415--447, 1992.

\bibitem[Mau12]{maurer2012thermodynamics}
Andreas Maurer.
\newblock Thermodynamics and concentration.
\newblock {\em Bernoulli}, 18(2):434--454, 2012.

\bibitem[MF95]{merhav1995strong}
Neri Merhav and Meir Feder.
\newblock A strong version of the redundancy-capacity theorem of universal coding.
\newblock {\em IEEE Transactions on Information Theory}, 41(3):714--722, 1995.

\bibitem[MM09]{mezard2009information}
Marc Mezard and Andrea Montanari.
\newblock {\em Information, physics, and computation}.
\newblock Oxford University Press, 2009.

\bibitem[MMS{\etalchar{+}}18]{madry2018towards}
Aleksander Madry, Aleksandar Makelov, Ludwig Schmidt, Dimitris Tsipras, and Adrian Vladu.
\newblock Towards deep learning models resistant to adversarial attacks.
\newblock In {\em International Conference on Learning Representations}, 2018.

\bibitem[MSS19]{mohri2019agnostic}
Mehryar Mohri, Gary Sivek, and Ananda~Theertha Suresh.
\newblock Agnostic federated learning.
\newblock In {\em International conference on machine learning}, pages 4615--4625. PMLR, 2019.

\bibitem[Mur12]{murphy2012machine}
Kevin~P Murphy.
\newblock {\em Machine learning: a probabilistic perspective}.
\newblock MIT press, 2012.

\bibitem[Ney35]{neyman1935suunteorema}
Jerzy Neyman.
\newblock Su un teorema concernente le cosiddette statistiche sufficienti.
\newblock {\em Giornale dell'Istituto Italiano degli Attuari}, VI(4):320--334, 1935.

\bibitem[NP36]{neyman1936sufficient}
Jerzy Neyman and Egon~S Pearson.
\newblock Sufficient statistics and uniformly most powerful tests of statistical hypotheses.
\newblock {\em Statistical Research Memoirs}, 1:113--137, 1936.

\bibitem[NZB17]{nguyen2017inverse}
H~Chau Nguyen, Riccardo Zecchina, and Johannes Berg.
\newblock Inverse statistical problems: from the inverse ising problem to data science.
\newblock {\em Advances in Physics}, 66(3):197--261, 2017.

\bibitem[Pat17]{pathria2017statistical}
Raj~Kumar Pathria.
\newblock {\em Statistical Mechanics: International Series of Monographs in Natural Philosophy}, volume~45.
\newblock Elsevier, 2017.

\bibitem[Pit36]{pitman1936sufficient}
Edwin James~George Pitman.
\newblock Sufficient statistics and intrinsic accuracy.
\newblock In {\em Mathematical Proceedings of the cambridge Philosophical society}, volume~32, pages 567--579. Cambridge University Press, 1936.

\bibitem[Pol04]{Pollard04}
David Pollard.
\newblock Hammersley-clifford theorem for markov random fields, 2004.

\bibitem[PR17]{peliti2017einstein}
Luca Peliti and Ra{\'u}l Rechtman.
\newblock Einstein’s approach to statistical mechanics: the 1902--04 papers.
\newblock {\em Journal of statistical physics}, 167:1020--1038, 2017.

\bibitem[R{\'e}n61]{renyi1961measures}
Alfr{\'e}d R{\'e}nyi.
\newblock On measures of entropy and information.
\newblock In {\em Proceedings of the fourth Berkeley symposium on mathematical statistics and probability, volume 1: contributions to the theory of statistics}, volume~4, pages 547--562. University of California Press, 1961.

\bibitem[San57]{Sanov57}
Ivan~Nikolaevich Sanov.
\newblock On the probability of large deviations of random magnitudes.
\newblock {\em Matematicheskii Sbornik}, 84(1):11--44, 1957.

\bibitem[Sch48]{schrodinger1948statistical}
Erwin Schr{\"o}dinger.
\newblock {\em Statistical thermodynamics}.
\newblock Cambridge University Press, 1948.

\bibitem[SDBCD24]{shi2024diffusion}
Yuyang Shi, Valentin De~Bortoli, Andrew Campbell, and Arnaud Doucet.
\newblock Diffusion schr{\"o}dinger bridge matching.
\newblock {\em Advances in Neural Information Processing Systems}, 36, 2024.

\bibitem[SF12]{SF12}
Robert~E. Schapire and Yoav Freund.
\newblock {\em Boosting: Foundations and Algorithms}.
\newblock The MIT Press, 2012.

\bibitem[Sha48]{shannon1948}
Claude~Elwood Shannon.
\newblock A mathematical theory of communication.
\newblock {\em The Bell system technical journal}, 27(3):379--423, 1948.

\bibitem[Sio58]{S58}
Maurice Sion.
\newblock On general minimax theorems.
\newblock {\em Pacific J. Math.}, 8(1):171--176, 1958.

\bibitem[SJ80]{shore1980axiomatic}
John Shore and Rodney Johnson.
\newblock Axiomatic derivation of the principle of maximum entropy and the principle of minimum cross-entropy.
\newblock {\em IEEE Transactions on information theory}, 26(1):26--37, 1980.

\bibitem[SJJ96]{saul1996mean}
Lawrence~K Saul, Tommi Jaakkola, and Michael~I Jordan.
\newblock Mean field theory for sigmoid belief networks.
\newblock {\em Journal of artificial intelligence research}, 4:61--76, 1996.

\bibitem[SKHL19]{sagawa2019distributionally}
Shiori Sagawa, Pang~Wei Koh, Tatsunori~B Hashimoto, and Percy Liang.
\newblock Distributionally robust neural networks for group shifts: On the importance of regularization for worst-case generalization.
\newblock {\em arXiv preprint arXiv:1911.08731}, 2019.

\bibitem[Ski88]{skilling1988axioms}
John Skilling.
\newblock The axioms of maximum entropy.
\newblock In {\em Maximum-Entropy and Bayesian Methods in Science and Engineering: Foundations}, pages 173--187. Springer, 1988.

\bibitem[SM15]{sharp2015translation}
Kim Sharp and Franz Matschinsky.
\newblock Translation of ludwig boltzmann’s paper “on the relationship between the second fundamental theorem of the mechanical theory of heat and probability calculations regarding the conditions for thermal equilibrium” sitzungberichte der kaiserlichen akademie der wissenschaften. mathematisch-naturwissen classe. abt. ii, lxxvi 1877, pp 373-435 (wien. ber. 1877, 76: 373-435). reprinted in wiss. abhandlungen, vol. ii, reprint 42, p. 164-223, barth, leipzig, 1909.
\newblock {\em Entropy}, 17(4):1971--2009, 2015.

\bibitem[SST92]{seung1992statistical}
Hyunjune~Sebastian Seung, Haim Sompolinsky, and Naftali Tishby.
\newblock Statistical mechanics of learning from examples.
\newblock {\em Physical review A}, 45(8):6056, 1992.

\bibitem[Tju74]{tjur1974conditional}
Tue Tjur.
\newblock {\em Conditional probability distributions}.
\newblock Institute of Mathematical Statistics, University of Copenhagen, 1974.

\bibitem[Tol79]{tolman1979principles}
Richard~Chace Tolman.
\newblock {\em The principles of statistical mechanics}.
\newblock Courier Corporation, 1979.

\bibitem[Top79]{topsoe1979information}
Flemming Tops{\o}e.
\newblock Information-theoretical optimization techniques.
\newblock {\em Kybernetika}, 15(1):8--27, 1979.

\bibitem[Top93]{topsoe1993game}
Flemming Tops{\o}e.
\newblock Game theoretical equilibrium, maximum entropy and minimum information discrimination.
\newblock In {\em Maximum Entropy and Bayesian Methods: Paris, France, 1992}, pages 15--23. Springer, 1993.

\bibitem[Tou09]{touchette2009large}
Hugo Touchette.
\newblock The large deviation approach to statistical mechanics.
\newblock {\em Physics Reports}, 478(1-3):1--69, 2009.

\bibitem[TTL84]{tikochinsky1984alternative}
Y~Tikochinsky, NZ~Tishby, and Raphael~David Levine.
\newblock Alternative approach to maximum-entropy inference.
\newblock {\em Physical Review A}, 30(5):2638, 1984.

\bibitem[VCC81]{van1981maximum}
Jan Van~Campenhout and T~Cover.
\newblock Maximum entropy and conditional probability.
\newblock {\em IEEE Transactions on Information Theory}, 27(4):483--489, 1981.

\bibitem[Wat09]{watanabe2009algebraic}
Sumio Watanabe.
\newblock {\em Algebraic geometry and statistical learning theory}, volume~25.
\newblock Cambridge university press, 2009.

\bibitem[WB98]{williams1998bayesian}
Christopher~KI Williams and David Barber.
\newblock Bayesian classification with gaussian processes.
\newblock {\em IEEE Transactions on pattern analysis and machine intelligence}, 20(12):1342--1351, 1998.

\bibitem[WJ{\etalchar{+}}08]{wainwright2008graphical}
Martin~J Wainwright, Michael~I Jordan, et~al.
\newblock Graphical models, exponential families, and variational inference.
\newblock {\em Foundations and Trends{\textregistered} in Machine Learning}, 1(1--2):1--305, 2008.

\bibitem[WT11]{welling2011bayesian}
Max Welling and Yee~W Teh.
\newblock Bayesian learning via stochastic gradient langevin dynamics.
\newblock In {\em Proceedings of the 28th international conference on machine learning (ICML-11)}, pages 681--688. Citeseer, 2011.

\bibitem[ZK16]{zdeborova2016statistical}
Lenka Zdeborov{\'a} and Florent Krzakala.
\newblock Statistical physics of inference: Thresholds and algorithms.
\newblock {\em Advances in Physics}, 65(5):453--552, 2016.

\end{thebibliography}

\newpage
\appendix

\end{document}